\documentclass[12pt, final]{l4dc2020}
\setcitestyle{numbers}
\setcitestyle{square}
\usepackage{subcaption}
\usepackage[export]{adjustbox}
\usepackage{wrapfig}

\newcommand{\algo}{SDS-EF\xspace}

\usepackage{amsmath}
\usepackage{amsfonts}
\usepackage{amssymb}
\usepackage{bm}
\usepackage{bbm}
\usepackage{mathtools}
\usepackage{enumitem}
\usepackage{thmtools,thm-restate}
\usepackage{algorithm}
\usepackage{algorithmic}
\usepackage{graphicx}
\usepackage{comment}

\def\MM{\mathcal{M}}

\def\Jb{\mathbf{J}}

\def\Rbb{\mathbb{R}}

\def\R{\Rbb}

\def\t{\top}
\def\*{\star}

\newcommand{\w}{\mathbf{w}}

\newcommand{\x}{\mathbf{x}}
\newcommand{\xd}{{\dot{\x}}}

\newcommand{\y}{\mathbf{y}}
\newcommand{\yd}{{\dot{\y}}}

\newcommand{\z}{\mathbf{z}}

\newcommand{\J}{\mathbf{J}}

\newcommand{\G}{\mathbf{G}}

\newcommand{\sdot}[2]{\overset{\lower0.1em\hbox{$\scriptscriptstyle #2$}}{#1}}

\title[Euclideanizing Flows: Diffeomorphic Reduction for Learning Stable Dynamical Systems]{Euclideanizing Flows: Diffeomorphic Reduction for Learning\\
Stable Dynamical Systems}
\usepackage{times}

\author{%
\Name{M. Asif Rana$^{1,4}$,  Anqi Li$^{2,4}$,   Dieter Fox$^{2,4}$,  Byron Boots$^{2,4}$,  Fabio Ramos$^{3,4}$,  Nathan Ratliff$^{4}$}
 \AND
 \addr $^1$Georgia Institute of Technology, \quad $^2$University of Washington,\quad $^3$University of Sydney, \quad $^4$NVIDIA%
}

\begin{document}

\maketitle
\vspace{-1em}
\begin{abstract}%
Robotic tasks often require motions with complex geometric structures. We present an approach to learn such motions from a limited number of human demonstrations by exploiting the regularity properties of human motions e.g. stability, smoothness, and boundedness. The complex motions are encoded as rollouts of a stable dynamical system, which, under a change of coordinates defined by a diffeomorphism, is equivalent to a simple, hand-specified dynamical system. As an immediate result of using diffeomorphisms, the stability property of the hand-specified dynamical system directly carry over to the learned dynamical system. 
Inspired by recent works in density estimation, we propose to represent the diffeomorphism as a composition of simple parameterized diffeomorphisms. Additional structure is imposed to provide guarantees on the smoothness of the generated motions. The efficacy of this approach is demonstrated through validation on an established benchmark as well demonstrations collected on a real-world robotic system.
\end{abstract}


\section{Introduction}
\vspace{-0.5em}


In many applications, robots are required to execute complex motions in potentially dynamic and unstructured environments. Since hand-coding such motions can be cumbersome or even infeasible, learning from demonstration (LfD)~\cite{argall2009survey} enables robots to, instead, acquire new motion skills by observing humans. In this work, we focus on learning goal-directed motions from human demonstrations, i.e. motions that stop at a given target location. This is without loss of generality since complex tasks can often be achieved by an ordered execution of goal-directed motions~\cite{schaal1999imitation}. 

Human motions naturally preserve regularity properties, e.g. continuity, smoothness and boundedness~\cite{flash1985coordination}. Since we assume goal-directed motions, human demonstrations should also naturally be stable. Exploiting these regularity properties can significantly improve the sample efficiency of the learning algorithms, which is critical since the number of human demonstrations are often limited. Thus, we model human motions as rollouts from a stable dynamical system. Motions governed by stable dynamical systems exhibit some additional desirable properties. First, such motions can react to temporal and spatial perturbations, which is necessary in dynamic task settings. Second, the stability property formally guarantees that the motions converge to the goal region. 

The challenge, however, is to encode stable human motions into dynamical systems that are stable by construction. A number of approaches have been proposed, which address this problem by explicitly parameterizing the class of stable dynamical systems and the notions of stability~\cite{khansari2011learning,khansari2014learning,ravichandar2017learning}. 
In this work, we take a fundamentally different approach than the aforementioned methods. Instead of explicitly learning a stable dynamical system, we view demonstrations as motions on a Riemannian manifold which is linked, under a smooth bijective map, i.e. a \emph{diffeomorphism}, to a latent Euclidean space. This diffeomorphism implicitly gives rise to an inherently stable dynamical system. The learning problem thus involves finding a diffeormorphism which explains the observed demonstrations. Compared to existing diffeomorphism learning approaches to encoding motions~\cite{neumann2015learning, perrin2016fast}, our formulation, \textbf{S}table \textbf{D}ynamical \textbf{S}ystem learning using \textbf{E}uclideanizing \textbf{F}lows (\algo), is based on a more expressive formulation of diffeomorphisms. 



We present an approach for learning a time-invariant continuous-time dynamical system (or reactive motion policy), which is globally asymptotically stable. Our dynamics formulation allows encoding severely curved goal-directed motions, and can be learned from a few demonstrations with minimal parameter tuning. Our specific contributions include: {(i)} a formulation of stable dynamics through warping curves into simple motions on a latent space using diffeomorphisms, and {(ii)} an expressive class of diffeomorphisms suitable for learning stable and smooth dynamical systems. We demonstrate the effectiveness of our approach on a standard handwriting dataset~\cite{khansari2019lasa}, and data collected on a robot manipulator~\cite{rana2019learning}.

\vspace{-0.5em}
\section{Related Work}
\vspace{-0.5em}

Over the past decade, a number of approaches have been proposed towards learning goal-directed time-invariant dynamical systems from human demonstrations~\cite{osa2018algorithmic}. An early approach is SEDS~\cite{khansari2011learning}. SEDS assumes the demonstrations comply with a Lyapunov (or Energy) function given by the squared-distance to the goal, and thus is restricted to motions which monotonically converge to the goal over time. A relaxation to the stability criterion was proposed in CLF-DM~\cite{khansari2014learning}, which instead assumes Lyapunov functions in the form of a weighted sum of asymmetric quadratic functions (WSAQF). The parameters of the Lyapunov function are learned independently from the (unstable) dynamics, and then used in an online fashion to generate stabilizing controls. The online correction scheme may interfere significantly with the learned dynamics~\cite{neumann2015learning}. A more recent approach, CDSP~\cite{chaandar2019learning} instead enforces incremental stability, a notion concerned with relative displacement of motions. Instead of learning a Lyapunov function, CDSP proposed learning a positive-definite contraction metric. However, CDSP restricts the class of contraction metrics to (potentially limiting) sum of squared polynomials.

There are a few approaches that learn stable dynamical systems via learned diffeomorphisms. 
However, by definition, a diffeomorphism must be invertible, making the learning problem non-trivial. One approach~\cite{perrin2016fast}, realizes a diffeomorphism as a composition of locally weighted translations. The authors only apply this approach to the problem of learning a single (or average) demonstration. Another strategy, $\tau$-SEDS~\cite{neumann2015learning}, learns diffeomorphisms from multiple demonstrations. However, $\tau$-SEDS defines a diffeomorphism by the square-root of a WSAQF, thus restricting the hypothesis class. In contrast to this prior work, we learn diffeomorphisms by using flexible function approximators  including kernel methods~\cite{scholkopf1999advances} and neural networks~\cite{lecun1998gradient}. 


Our diffeomorphism learning approach builds on normalizing flows~\cite{grathwohl2018ffjord,tabak2013family}, which have recently been successfully used for density estimation. The goal of normalizing flows is to map a simple base distribution into a complicated probability distribution over observed data by applying the change of variable theorem sequentially. Our problem is similar to the normalizing flows problem as we seek to map simple straight-line motions to complicated motions captured from human demonstrations. However, our problem is fundamentally different in two ways. First, normalizing flows only require the mapping to be bijective, a property less strict than diffeomorphism (see Section~\ref{sec:background}). Second, normalizing flows are concerned primarily with mapping scalar functions to scalar functions (i.e. probability densities). In our method, we seek to map vector fields to vector fields.

 \begin{figure}
 \centering
     \fbox{\subfigure[]{\includegraphics[trim=1.54cm 0.4cm 1.54cm 0.4cm, clip, width=0.234\textwidth]{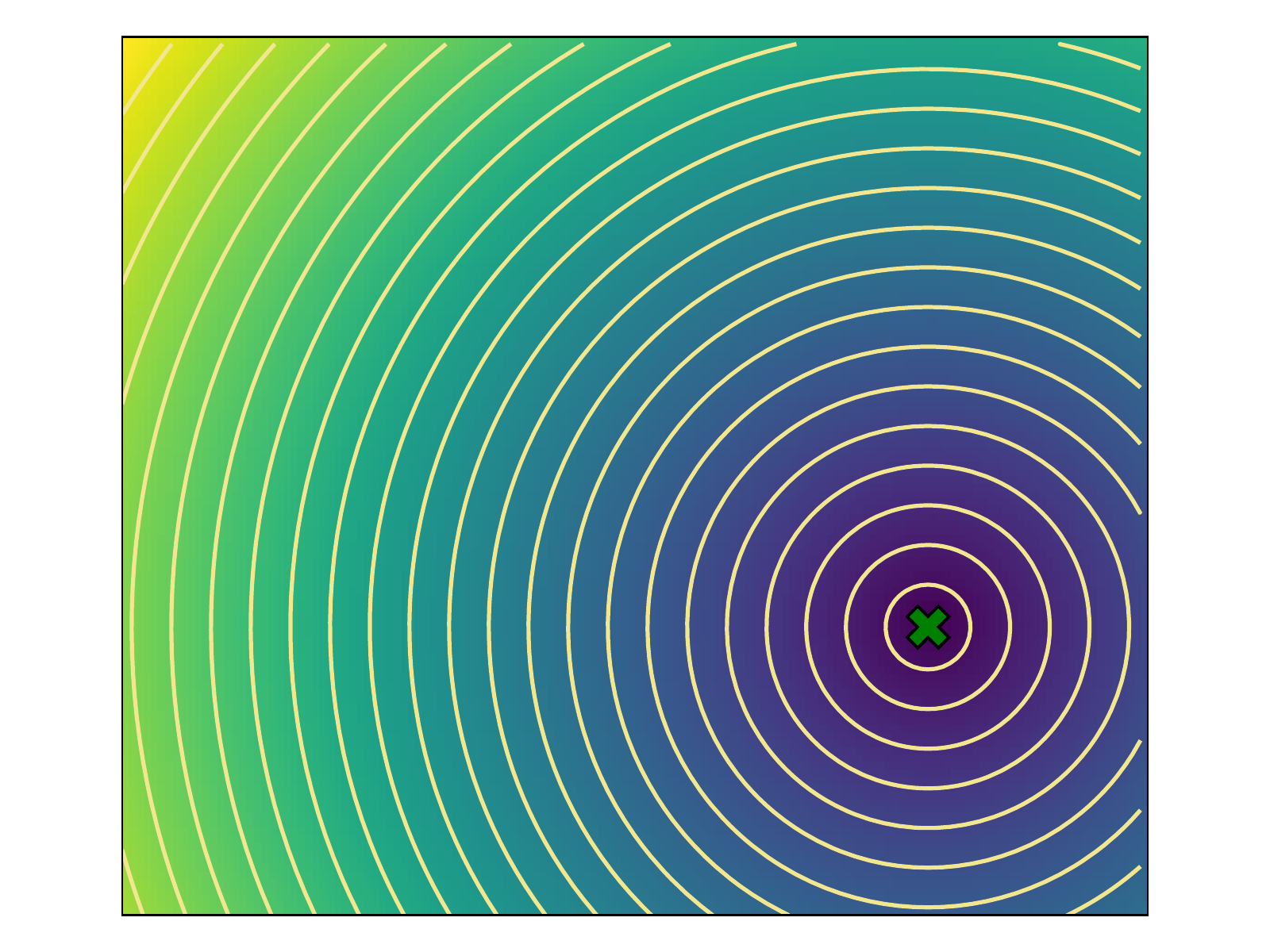}} 
     \subfigure[]{\includegraphics[trim=1.54cm 0.4cm 1.54cm 0.4cm, clip, width=0.234\textwidth]{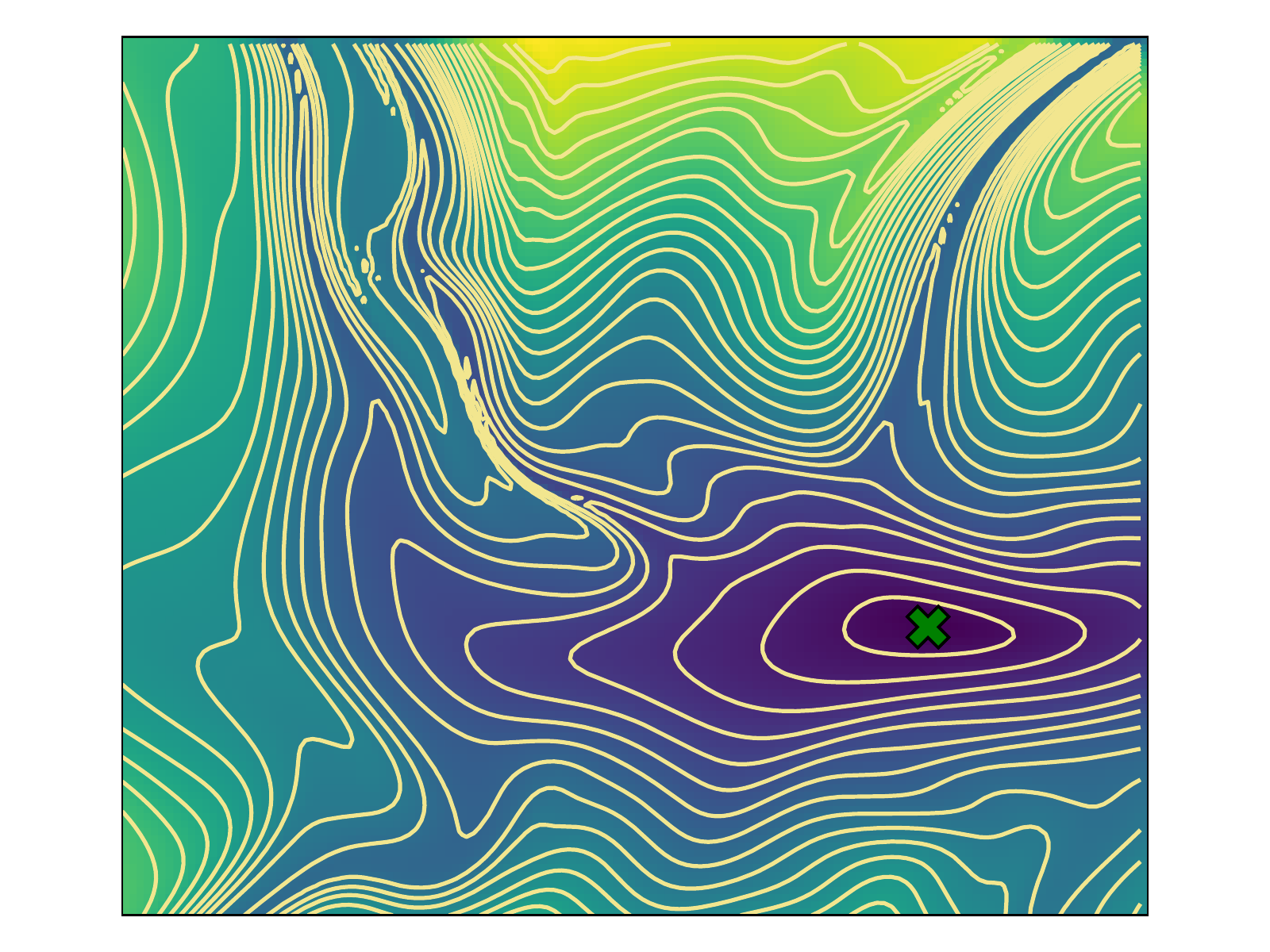}}}
     ~
     \fbox{\subfigure[]{\includegraphics[trim=1.54cm 0.4cm 1.54cm 0.4cm, clip, width=0.234\textwidth]{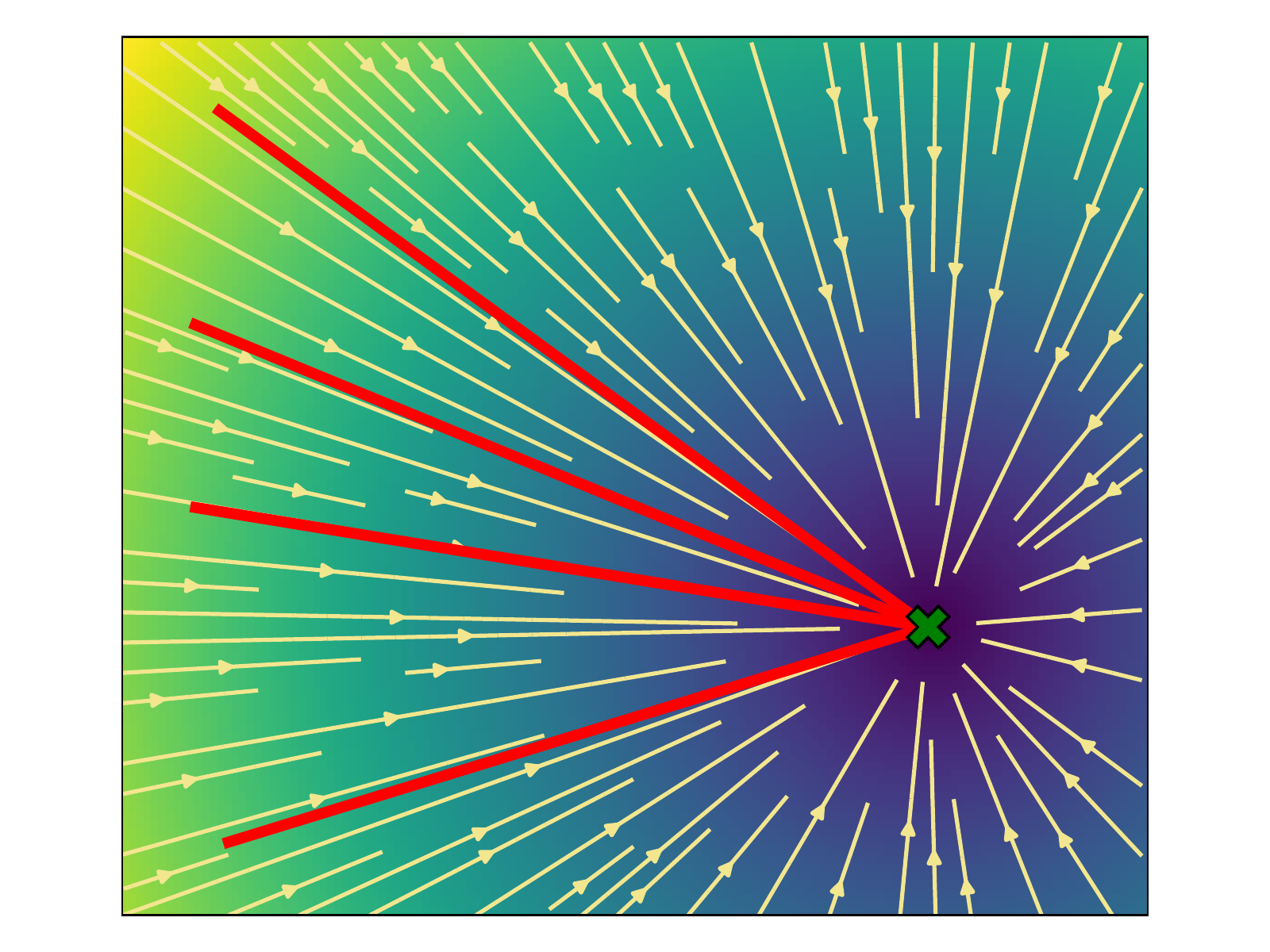}}
      \subfigure[]{\includegraphics[trim=1.54cm 0.4cm 1.54cm 0.4cm, clip, width=0.234\textwidth]{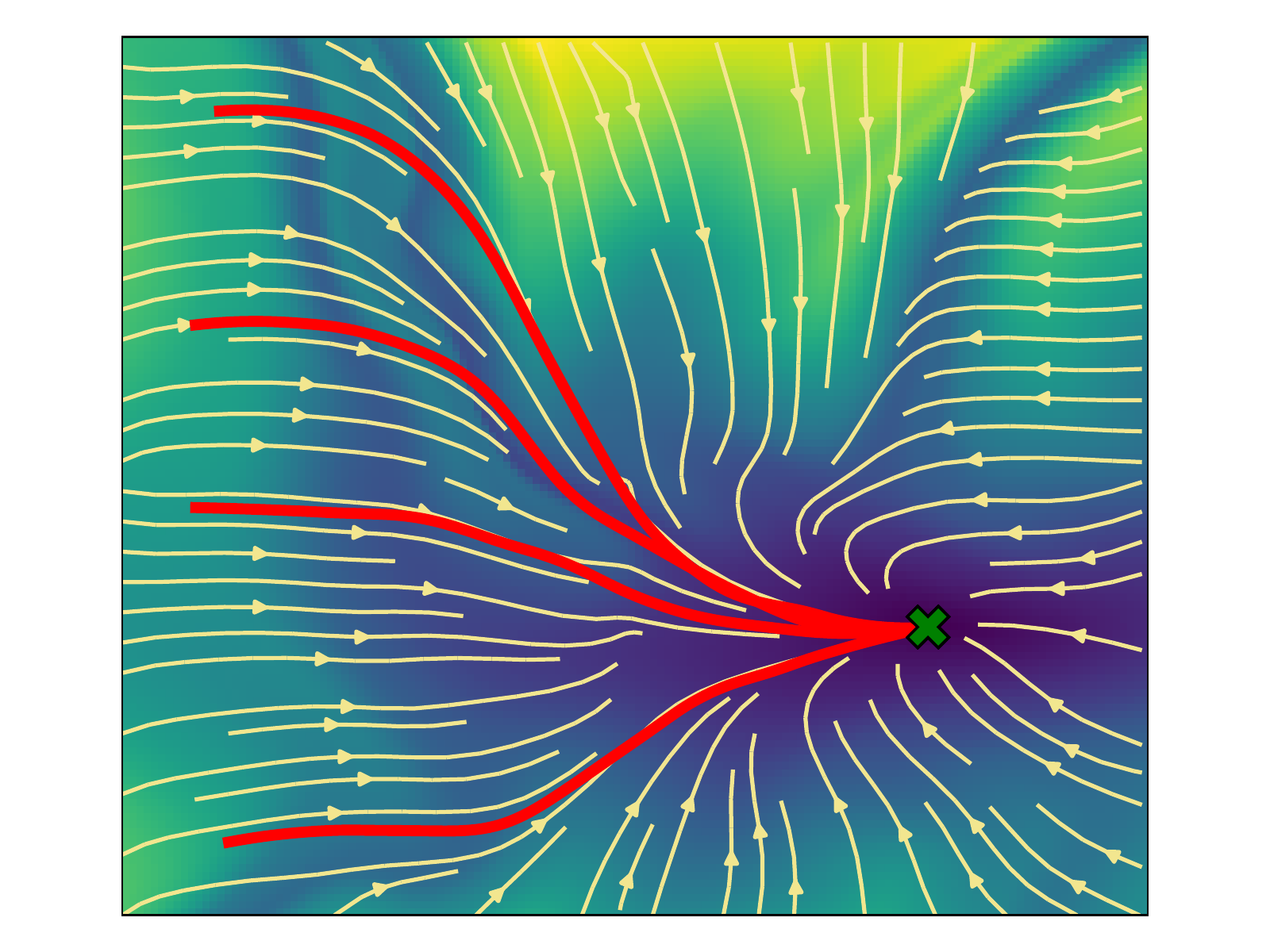}}}
\caption{\small{\emph{(a)}-\emph{(b)}: Isocontours showing equidistant points from the origin (green cross) in a Euclidean space and a Riemannian manifold respectively. \emph{(c)}-\emph{(d)}: Velocity fields governing shortest distance paths to the origin in the two spaces; rollouts from specific locations are overlayed in red.}}
\label{fig:grid_plots}
 \end{figure}

\vspace{-0.5em}
\section{Background: Stability, Diffeomorphism, and Riemannian Manifolds}\label{sec:background}
\vspace{-0.5em}
We briefly summarize the theoretical background for this paper. First, we introduce the concept of global asymptotic stability and how it can be shown through an auxiliary function, i.e. a Lyapunov function. Next, we discuss how one dynamical system can be described in different ways through a change of coordinates using a diffeomorphism. Finally, we shed light into the geometric interpretation of diffeomorphisms, especially when applied to gradient descent dynamical systems. 

\textbf{Notation: }To define mappings, we use symbol $\to$ to specify domains and co-domians, e.g. $\psi:\R^n\to\R$, and $\mapsto$ for how individual elements of the domains are mapped, e.g. $\psi:\x\to\y$. We use both symbols $\nabla$ and $\partial$ to denote derivatives, with a transposed relationship: consider $\x\in\R^n$ and a differentiable map $\psi: \R^n\to \R$, then $\nabla_\x \psi(\x)=\frac{\partial \psi}{\partial \x}^\t$. To evaluate functions which involve derivatives denoted by $\partial$, we use subscripts of brackets, e.g., $\big[\frac{\partial \psi}{\partial \x}\big]_{\x=0}=\nabla_\x \psi(0)^\t$. We use the notation $\circ$ for function composition, i.e., for $f:\R^m\to\R^p$ and $g:\R^n\to\R^m$, $(f\circ g) (\x) = f(g(\x))$. 



\vspace{-0.5em}
\subsection{Global Asymptotic Stability}
\vspace{-0.5em}
Consider an $n$-dimensional differentiable manifold $\MM$ with a global coordinate $\x:\MM\to\R^n$, which maps the manifold $\MM$ to a point in $\R^n$. With a slight abuse of notation, we will use the coordinate $\x$ to both represent the map $\x:\MM\to\R^n$ and the point $\x\in\R^n$. A dynamical system on the manifold $\MM$ can be described as a dynamical system on $\R^n$ under coordinate $\x$. We consider the dynamical system in the following form,\vspace{-0.5em}
\begin{equation}\label{eq:ds_x}
    \xd = f(\x),\quad\text{where $f:\R^n\to\R^n$ is locally Lipschitz continuous.}\vspace{-0.5em}
\end{equation}
While there are multiple definitions of stability, we are specifically concerned with whether the system can converge to a point of interest $\x^*\in\R^n$ from an arbitrary initial state on $\R^n$. Assume that $\x^*$ is an equilibrium point for the system, i.e., $f(\x^*)=0$, this desired convergent property can be characterized by \emph{global asymptotic stability}\footnote{We refer the reader to~\cite{khalil2002nonlinear} for a more rigorous and thorough introduction of stability properties of dynamical system.} of the equilibrium point $\x^*$. The global asymptotic stability can be shown through a continuously differentiable, positive-definite, and radially unbounded function called \emph{Lyapunov function}~\cite{khalil2002nonlinear}. A Lyapunov function $V:\R^n\to\R$ is a scalar valued function which satisfies that (i) $\dot{V}(\x)=\nabla_\x V(\x)^\t\,f(\x)<0$ for all $\x\neq\x^*$ and (ii) $\dot{V}(\x^*)=0$ at the equilibrium $\x^*$. If there exists such a Lyapunov function $V$, the equilibrium point $\x^*$ is globally asymptotically stable: all trajectories converge to the point $\x^*$. 

\vspace{-0.5em}
\subsection{Change of Coordinates for Dynamical Systems}
\vspace{-0.5em}
Consider a bijective map $\psi:\mathbb{R}^n\to\mathbb{R}^n$. The bijective map $\psi$ is a \emph{diffeomorphism} if both the map $\psi$ and its inverse map $\psi^{-1}$ are continuously differentiable. We further assume that the diffeomorphism $\psi$ is bounded, namely that it maps bounded vectors to bounded vectors. A diffeomorphism can be used to describe a \emph{change of coordinates} for differentiable manifolds. Consider again the manifold $\MM$ with global coordinates $\x\in\R^n$. A diffeomorphism $\psi:\x\mapsto \y$ creates another global coordinate $\y:\MM\to\R^n$ for the manifold $\MM$ through the map $\y=\psi(\x)$. 
As such, the dynamical system~\eqref{eq:ds_x} can be described under the other coordinate $\y$ as,\vspace{-0.5em}
\begin{equation}\label{eq:ds_y}
    \yd = \left[\frac{\partial \psi}{\partial \x}\,f(\x)\right]_{\x=\psi^{-1}(\y)}= \J_\psi(\psi^{-1}(\y))\,f(\psi^{-1}(\y)) \quad:= \tilde{f}(\y),\vspace{-1.0em}
\end{equation}
where $\Jb_{\psi}(\x) = \frac{\partial \psi}{\partial \x}$ is the \emph{Jacobian matrix} of $\psi$. In the special case where both the system dynamics $f$ and the diffeomorphism $\psi$ are linear maps, this change of coordinates reduces to \emph{change of basis} for linear dynamical systems~\cite{hespanha2018linear}. 

The two dynamical systems~\eqref{eq:ds_x} and~\eqref{eq:ds_y} are descriptions of \emph{the same internal dynamical system} evolving on the manifold $\MM$. Therefore, the two systems share stability properties. Assume the existence of a Lyapunov function $V(\x)$ showing that the equilibrium point $\x^*$ is globally asymptotically stable. Then, the diffeomorphism $\psi$ defines a Lyapunov function $\tilde{V}:\y\mapsto V(\psi^{-1}(\y))$,\vspace{-0.5em}
\begin{equation}\small
    \begin{split}
            \dot{\tilde{V}}(\y) &= \left[\frac{\partial V}{\partial \x}\,\frac{\partial \psi^{-1}}{\partial \y}\,\yd\right]_{\x=\psi^{-1}(\y)} = \left[\frac{\partial V}{\partial \x}\,\left(\J_\psi(\x)\right)^{-1}\,\tilde{f}(\y)\right]_{\x=\psi^{-1}(\y)}\\ 
            &= \left[\frac{\partial V}{\partial \x}\,\left(\J_\psi(\x)\right)^{-1}\J_\psi(\x)\,f(\x)\right]_{\x=\psi^{-1}(\y)}
            =\left[\frac{\partial V}{\partial \x}\,\xd\right]_{x=\psi^{-1}(\y)}=\dot{V}(\psi^{-1}(\y)),
    \end{split}
\end{equation}
where the second equality follows from the implicit function theorem. 
Therefore, the system after the change of coordinate~\eqref{eq:ds_y} has a globally asymptotically stable equilibrium point $\y^*=\psi(\x^*)$. Moreover, since the diffeomorphism is bijective, the converse is also true: if there exists a Lyapunov function $\tilde{V}$ under coordinate $\y$, equilibrium point $\x^*$ is globally asymptotically stable. 

\vspace{-0.5em}
\subsection{Riemannian Manifolds and Natural Gradient Descent}\label{sec:natural_gd}
\vspace{-0.5em}
From a geometric standpoint, a diffeomorphism $\psi:\x\mapsto\y$ can also help us understand the geometry of a Riemannian manifold in relation to Euclidean geometry. The geometry on an $n$-dimensional Riemmanian manifold can be defined by a Riemannian metric $\G_{\psi}:\R^n\to\R_{++}^{n\times n}$, which gives a notion of distance on the manifold~\cite{lee2006riemannian}. 

To elaborate, let us define Euclidean goemetry on the co-domain, with coordinates $\y$. Consider a gradient descent dynamical system $\yd=\tilde{f}(\y)=-\nabla_\y\Phi(\y)$ with a potential function $\Phi:\R^n\to\R$. Then, the change of coordinates defined by the diffeomorphism $\psi$ provides a description of the same dynamics under $\x$ coordinate,\vspace{-0.5em}
\begin{equation}\label{eq:natural_gd}
     \xd\,=-\,\G_\psi(\x)^{-1}\,\nabla_\x\Phi(\psi(\x)) \quad := f_\psi(\x).\vspace{-0.5em}
\end{equation}
where the induced Riemannian metric in the domain is given by $\G_\psi(\x) = \J_\psi(\x)^\t\J_\psi(\x)\in \R_{++}^{n\times n}$. 
The aforementioned dynamics is known as \emph{natural gradient descent}, which is steepest descent on a Riemannian manifold~\cite{amari1998natural}, with respect to the potential function $\Phi\circ\psi$. The system~\eqref{eq:natural_gd} can generate sophisticated trajectories although the potential function $\Phi$ may only take a simple form. Moreover, if the potential function $\Phi$ is also positive definite,\footnote{$\Phi$ is strictly positive everywhere except $\y^*$, and $\Phi(\y^*)=0$} convex, continuously differentiable, and radially unbounded, then the potential function $\Phi$ is a valid Lyapunov function for the gradient descent system. Therefore, both the gradient descent system and natural gradient descent system~\eqref{eq:natural_gd} admit a globally asymptotically stable equilibrium point. 

Figure~\ref{fig:grid_plots}(a)--(b) shows the iso-contours of a potential function $\Phi$ which generate straight-line motions on the Euclidean space, and the corresponding potential function $\Phi\circ \psi$ on the Riemannian manifold. The isocontours show equidistant points to the goal in the Riemannian manifold, thus revealing the underlying geometry. Figure \ref{fig:grid_plots}(c)--(d) shows the trajectories generated by the same underlying system observed in the corresponding coordinate systems. 

\vspace{-0.5em}
\section{Learning Stable Dynamics Using Diffeomorphisms}
\vspace{-0.5em}
We view human demonstrations as goal-directed motions on an $n$-dimensional Riemannian manifold, governed by a stable dynamical system of the form \eqref{eq:natural_gd}. We view this dynamical system to be equivalent, under a change of coordinates, to another system defined on a \emph{latent} space. Our key insight here is: \emph{a diffeomorphism can warp a simple potential function into a more complicated one, and hence transform straight-line trajectories into severely curved motions.}  As a result, the problem of learning stable dynamical systems reduces to a diffeomorphism learning problem.  

\vspace{-0.5em}
\subsection{Problem Statement}
\vspace{-0.5em}
Assume the availability of $N$ human demonstrations, each composed of $T_i$ position-velocity pairs, denoted by $\{\{(\tilde{\x}_{i,t}, \dot{\tilde{\x}}_{i,t})\}_{t=1}^{T_i}\}_{i=1}^{N}$. We seek to find a dynamical system $\dot{\x} = f_{\psi}(\x)$ that reproduces the demonstrations while ensuring stability. As illustrated in Section \ref{sec:natural_gd}, we represent the system as the gradient descent system $\yd = -\nabla_\y\Phi(\y)$ after a change of coordinates defined by $\psi$. 

Although both the diffeomorphism $\psi$ and the potential function $\Phi$ can shape the dynamical system~\eqref{eq:natural_gd}, we view them as playing fundamentally different roles: the potential function $\Phi$ dictates the theoretical property of the dynamics, e.g. stability guarantees, while the diffeomorphism $\psi$ provides expressivity to our hypothesis class. As a result, we specify a simple potential function $\Phi(\y)=\|\y -\y^*\|$, with $\y^*=\psi(\x^*)$. This potential function generates unit-velocity straight-line motions to the globally asymptotically stable equilibrium point $\y^*$. The diffeomorphism acts to deform these straight lines to arbitrarily curved motions converging to $\x^*$, where the demonstrations converge. With a parameterized diffeomorphism $\psi_\theta$, the learning problem reduces to solving,\vspace{-0.5em}
\begin{equation}\label{eq:learning_prob}\small
    \hat{\theta} = \underset{\theta}{\arg\min}\,\frac{1}{\sum_{i=1}^N T_i}\sum_{i=1}^N\sum_{t=1}^{T_i}\,\big\|\dot{\tilde{\x}}_{i,t} - f_{\psi_\theta}({\tilde{\x}}_{i,t})\big\|_2^2\vspace{-0.5em}
\end{equation}
To avoid notation complexity, we will drop the subscript $\theta$ in the remainder of this paper.

\begin{figure}
    \vspace{-8mm}
    \centering
    \includegraphics[width=0.8\textwidth]{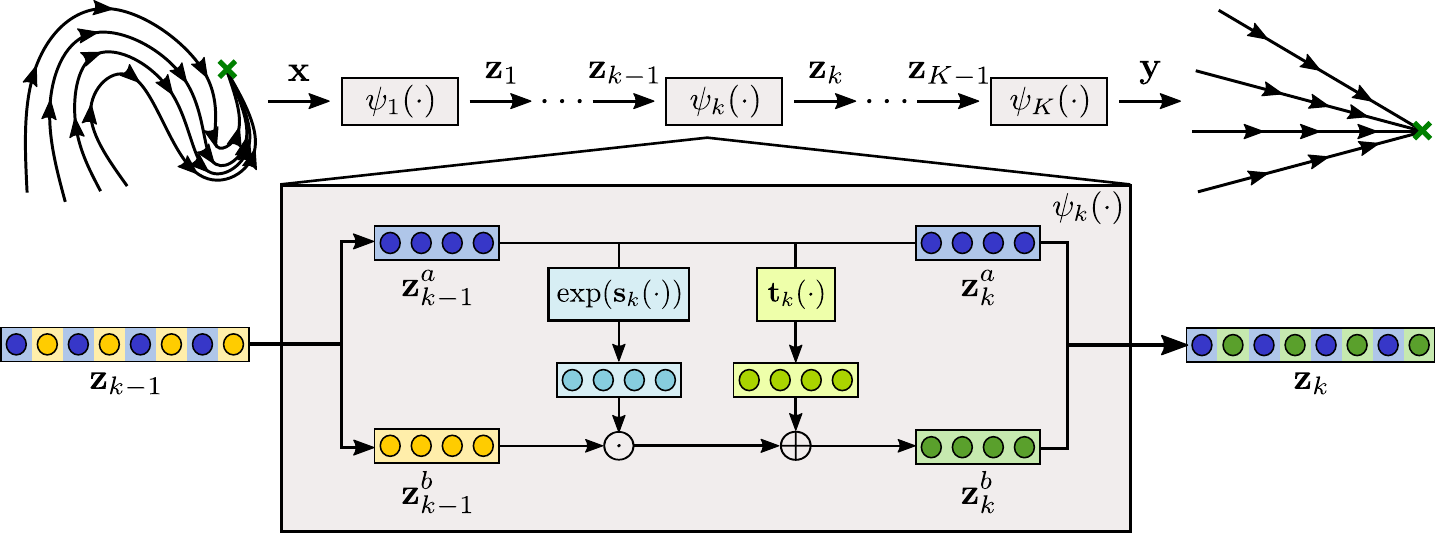}
    \vspace{4mm}
    \caption{Architecture of the diffeomorphic mapping network.\vspace{2mm}}
    \label{fig:diffeomorphism_net}
\end{figure}
\vspace{-0.5em}
\subsection{A Class of Expressive Diffeormorphisms}\label{sec:diffeomorphisms}
\vspace{-0.5em}
By definition, a diffeomorphism is required to be both, \emph{bijective} and \emph{continuously differentiable}. To achieve this, we impose structure in the learning problem by realizing a diffeomorphism by a composition of $K$ diffeomorphisms $\psi = \psi_1\circ\psi_2\circ\dots\circ\psi_K$, with each diffeomorphism $\psi_k:\R^n\to\R^n$ given by a \emph{coupling layer}~\cite{dinh2016density}. Let $\z_{k}\in\R^n$ denote the output from the $k$th coupling layer $\psi_{k-1}$, i.e., $\z_k=\psi_k(\z_k)$, $\z_0=\x$, and $\z_K=\y$. Each coupling layer, $\psi_k$ involves splitting the input $\z_{k-1}$ followed by scaling and translating one of the parts. The \texttt{split} operation divides inputs into two parts, $\z_{k-1}^a\in\R^{\lfloor n/2\rfloor}$ and $\z_{k-1}^b\in\R^{\lceil n/2 \rceil}$, constituting alternate input dimensions, whereby the pattern of alternation is reversed after each layer. Formally,\vspace{-0.5em}

\begin{equation}\label{eq:coupling_layer}\small
    \z_k = 
    \begin{bmatrix}
    \z_k^a\\\z_k^b
    \end{bmatrix} = 
    \begin{bmatrix}
    \z_{k-1}^a\\
    \z_{k-1}^b \odot \exp \big( {s}_k(\z_{k-1}^{a}) \big) + {t}_k(\z_{k-1}^{a})
    \end{bmatrix} \coloneqq \psi_k(\z_{k-1}),\vspace{-0.5em}
\end{equation}
where $\odot$ denotes pointwise product and $\exp$ denotes pointwise exponential. The functions ${s}_k:\R^{\lfloor n/2\rfloor}\to\R^{\lceil n/2\rceil}$ and ${t}_k:\R^{\lfloor n/2\rfloor}\to\R^{\lceil n/2\rceil }$ are the parameterized scaling and translation functions, respectively. The {coupling} layer is a bijective affine mapping by construction. Since each mapping is bijective, the composed mapping $\psi$ is also guaranteed to be bijective. Further, the composed mapping $\psi$ is continuously differentiable as long as the scaling and translation functions in each coupling layer are continuously differentiable. This formulation of a bijection has been previously employed for density estimation by Dinh et al. \cite{dinh2016density}, where the scaling and translation functions were given by deep convolutional neural networks.

In contrast to~\cite{dinh2016density}, for the scaling and translation functions, we use single-layer neural networks with the layer resembling an approximated kernel machine~\cite{rahimi2008random}. This special network structure leverage the advantages of both kernel machines and neural networks: (i) the kernel functions act as regularizers to enforce desired properties on the learned function, e.g. smoothness, which can further improve the sample efficiency of the learning algorithm, and (ii) as a parameterized model, the composed diffeomorphism can be efficiently trained using learning techniques for neural networks. 

We use a matrix-valued Gaussian separable kernel with length-scale $l$, defined as $K(\z,\z') = \exp (-\frac{\|\z-\z'\|^2}{2l^{2}})\mathbf{I}$. The Gaussian kernel restricts the hypothesis class to the class of $C^{\infty}$ vector-valued functions, imposing a stricter smoothness constraint than continuous differentiability, i.e. $C^{1}$. This is desirable since human motions are known to be maximizing smoothness (or minimizing jerk)~\cite{flash1985coordination}. We approximate the aforementioned kernel by $m$ randomly sampled Fourier features~\cite{rahimi2008random}, such that the scaling and translation functions are given by linear combinations of these features,\vspace{-0.7em}
\begin{align}\label{eq:rff}\small
    g(\z) = \varphi(\z)^\t\w,\ \ \ \ \ \ \ \  \text{with }
    \varphi(\z) = \sqrt{\frac{2}{m}}
    \begin{bmatrix}
    \cos(\bm{\alpha}_1^\t\z + \bm{\beta}_1)\\
    \cos(\bm{\alpha}_2^\t\z + \bm{\beta}_2)\\
    \vdots\\
    \cos(\bm{\alpha}_m^\t\z + \bm{\beta}_m)
    \end{bmatrix} \otimes \mathbf{I} \in \mathbb{R}^{ (m\cdot \lceil n/2\rceil)\times \lceil n/2\rceil},
\end{align}\\[-1.2em]
where $\w \in \mathbb{R}^{(m\cdot \lceil n/2\rceil)}$ constitutes the learnable parameters. The projection vector $\varphi(\z)$ is composed of $m$ randomly sampled Fourier features such that the kernel matrix is given by $K(\z,\z') \approx \varphi(\z)^{\t} \varphi(\z')$. Concretely, the coefficients $\{\bm{\alpha}_i\}_{i=1}^m$ are sampled from a zero-mean Gaussian distribution $\mathcal{N}(\mathbf{0}, l^{-2}\mathbf{I})$, and the bias terms $\{\bm{\beta}_i\}_{i=1}^m$ sampled from a uniform distribution $\text{U}(0, 2\pi)$. Under the formulation in \eqref{eq:rff}, we define ${s}_k(\cdot)=\varphi(\cdot)^\t\w_{{s}_k}$ and ${t}_k(\cdot)=\varphi(\cdot)^\t\w_{{t}_k}$. The set of parameters in the diffeomorphism learning problem in \eqref{eq:learning_prob} is therefore given by $\theta \coloneqq \{\w_{{s}_k}, \w_{{t}_k}\}_{k=1}^{K}$. 

\vspace{-0.5em}
\subsection{Practical Considerations}
\vspace{-0.5em}
Regarding the choice of potential functions, there are a few alternatives, including various soft versions of $\ell_2$-norm, and quadratic potential functions, and other norms. In this paper, we view the potential function as dictating the stability properties while the diffeomorphisms provides expressivity to our hypothesis class. We tested the proposed $\ell_2$-norm, a soft version of $\ell_2$-norm, and also a quadratic potential function, and we did not observe significant differences in performance. This observation also shows that our parameterization of diffeomorphsim is expressive enough so that the choice of potential function is not significant as long as it provides desirable stability guarantees. 

To learn~\eqref{eq:learning_prob} through back-propogation, the Jacobian of the diffeomorphism can be calculated analytically from~\eqref{eq:coupling_layer} and~\eqref{eq:rff}, or through auto-differentiation packages, e.g. pytorch~\cite{paszke2017automatic}.

\vspace{-0.6em}
\section{Experimental Results}
\vspace{-0.5em}
We evaluate our approach on the LASA dataset~\cite{khansari2019lasa} as well data collected for multiple tasks on a Franka Emika robot~\cite{rana2019learning}. For all our experiments, we start the learning procedure by normalizing the demonstrations to stay in the range $[-0.5, 0.5]$. This allows us to fix the hyperparameters of our model irrespective of the scale of the data. In our experiments, we use $K=10$ coupling layers constituting of $m=200$ random Fourier features with length-scale $l=0.45$. For consistency in results, the diffeomorphism is always initialized with an identity map. We optimize using ADAM~\cite{kingma2014adam} with default hyperparameters, alongside a learning rate of  $1\times 10^{-4}$ and an $\ell_2$-regularization on the weights $\theta$ with coefficient $1\times 10^{-8}$. 

\begin{figure}[htb]
\vspace{-2mm}
\begin{tabular}{cccccc}
\includegraphics[trim=2.5cm 0.1cm 2.5cm 0cm, clip, width = 0.16\linewidth]{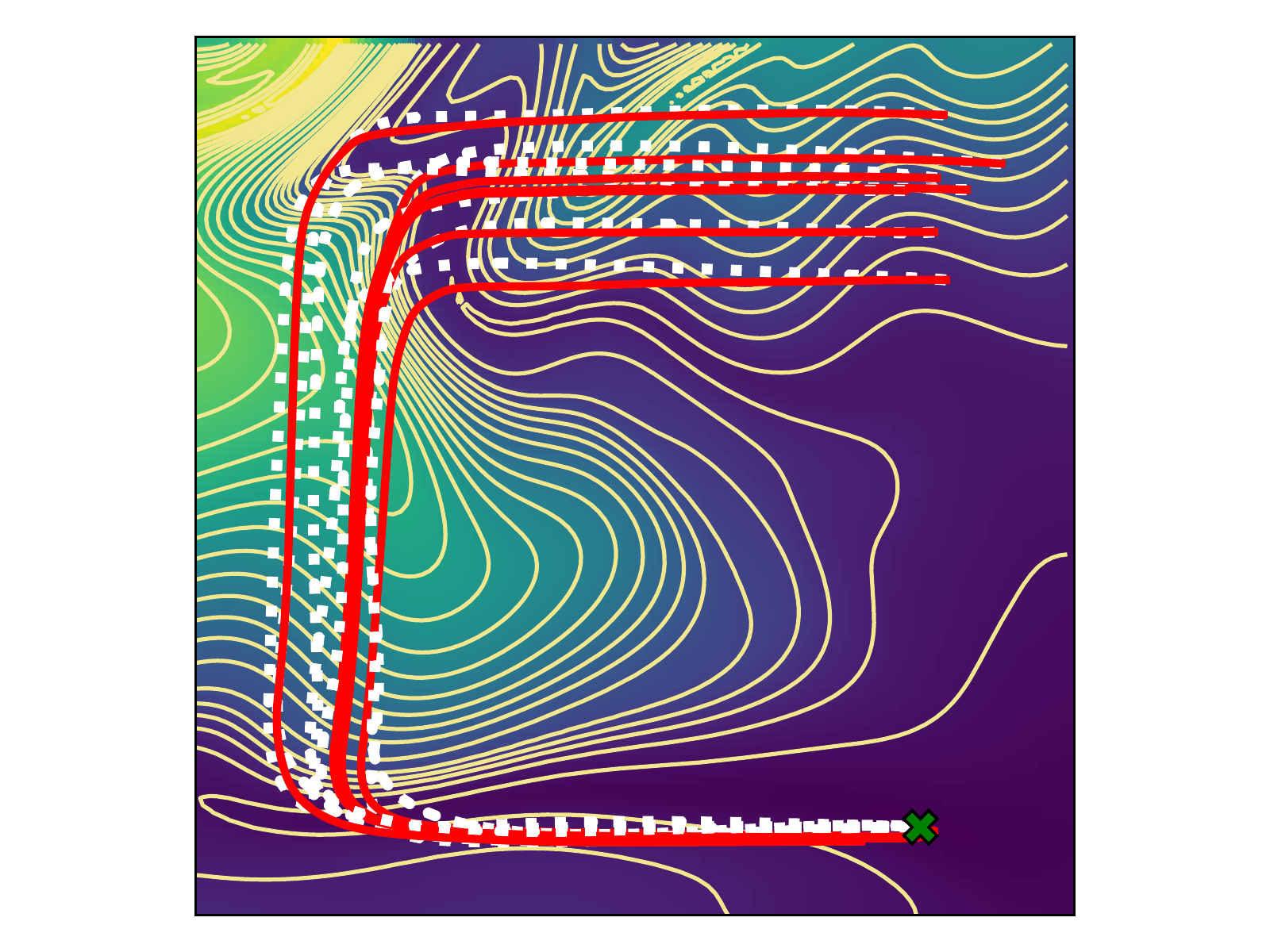}\hspace{-0.05cm}
\includegraphics[trim=2.5cm 0.1cm 2.5cm 0cm, clip, width = 0.16\linewidth]{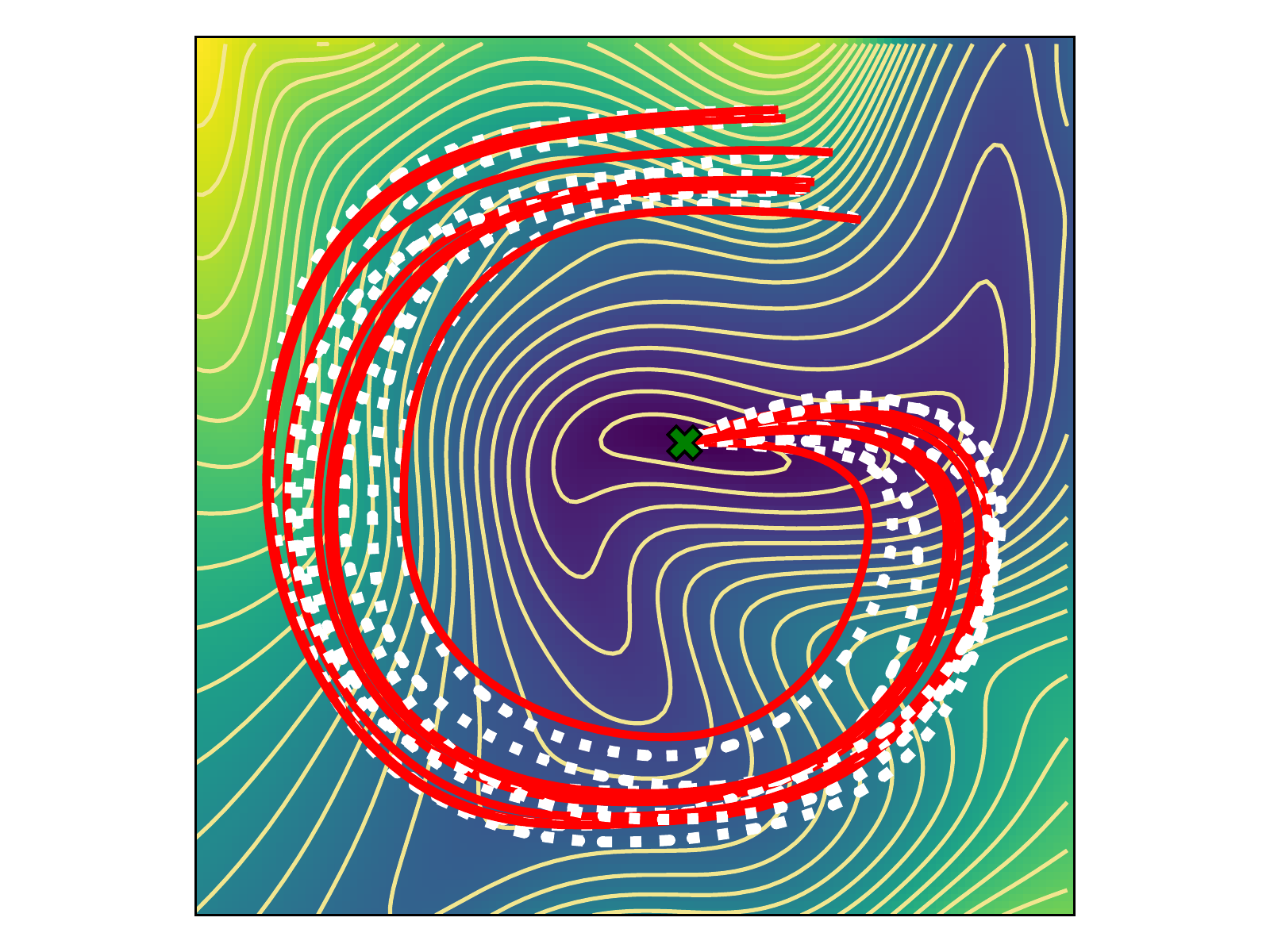}\hspace{-0.05cm}
\includegraphics[trim=2.5cm 0.1cm 2.5cm 0cm, clip, width = 0.16\linewidth]{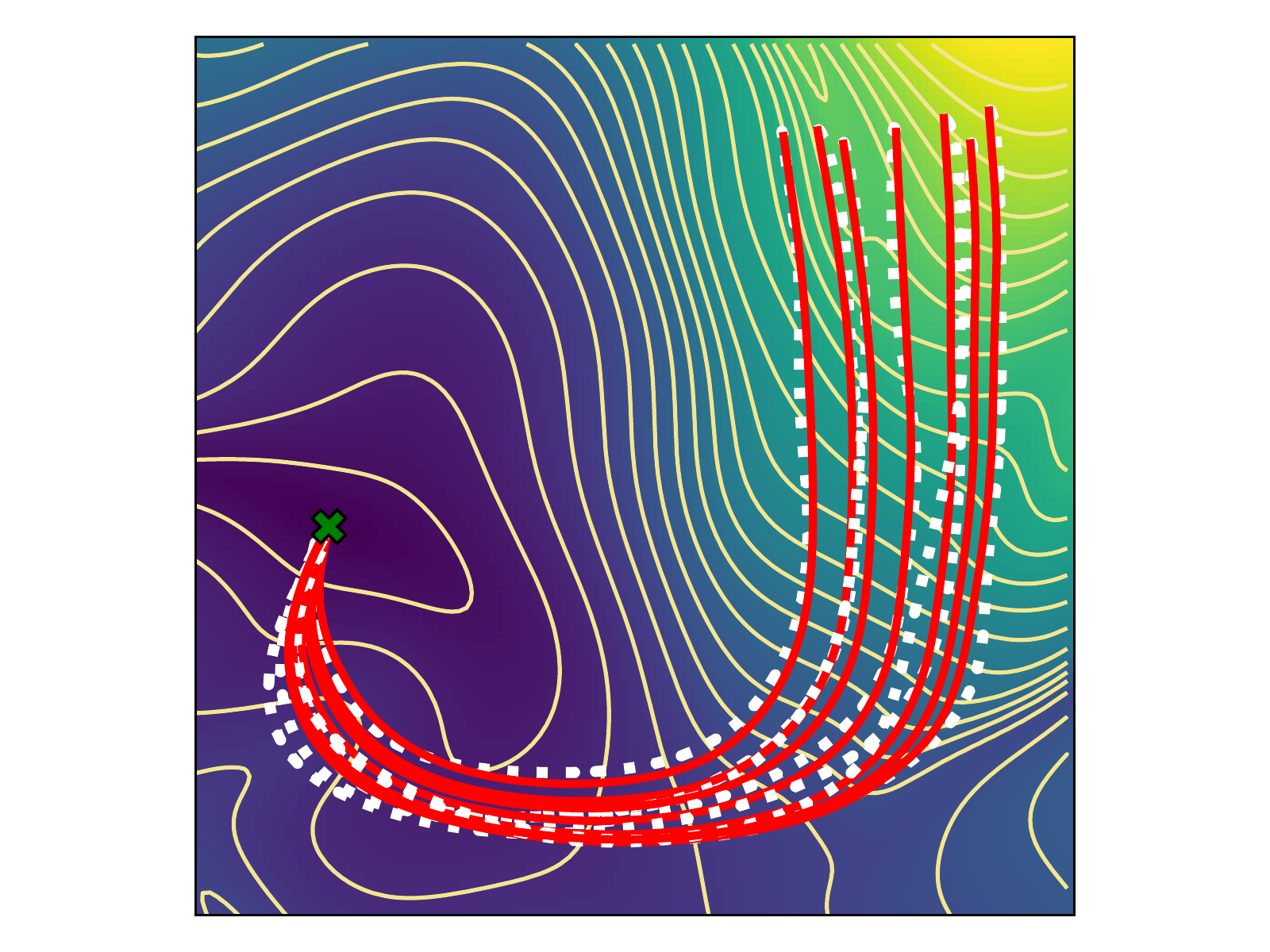}\hspace{-0.05cm} 
\includegraphics[trim=2.5cm 0.1cm 2.5cm 0cm, clip, width = 0.16\linewidth]{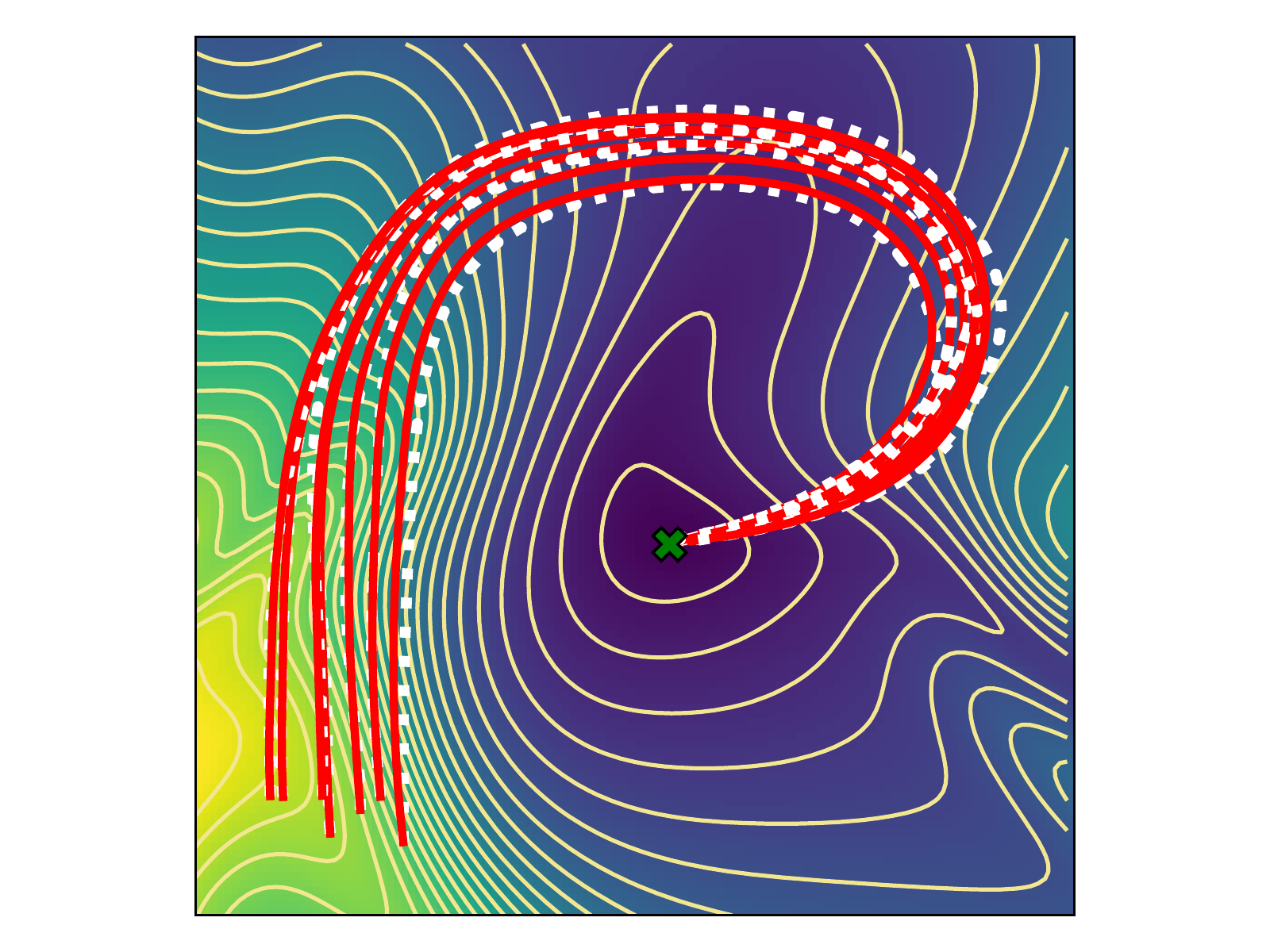}\hspace{-0.05cm}
\includegraphics[trim=2.5cm 0.1cm 2.5cm 0cm, clip, width = 0.16\linewidth]{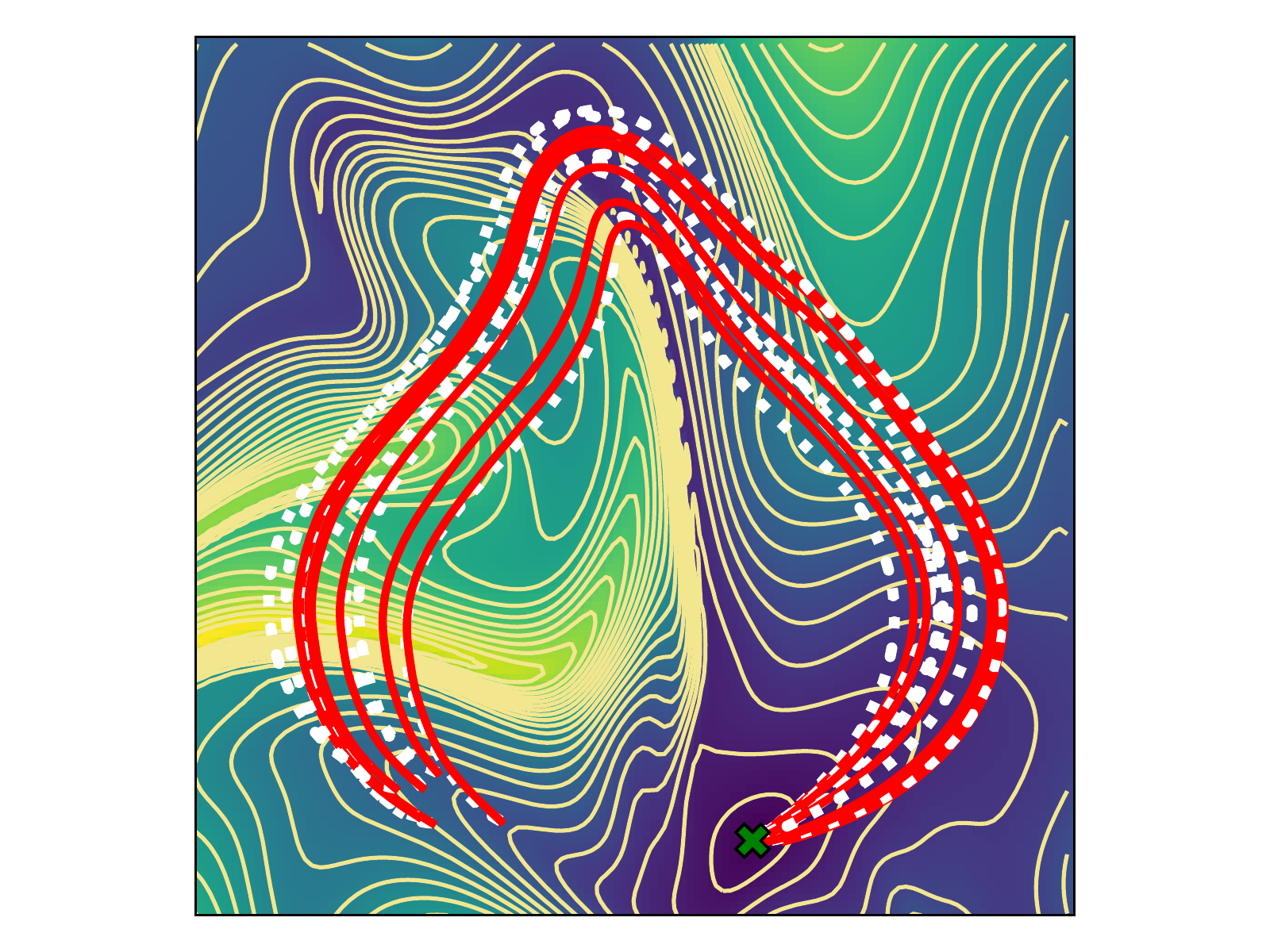}\hspace{-0.05cm}
\includegraphics[trim=2.5cm 0.1cm 2.5cm 0cm, clip, width = 0.16\linewidth]{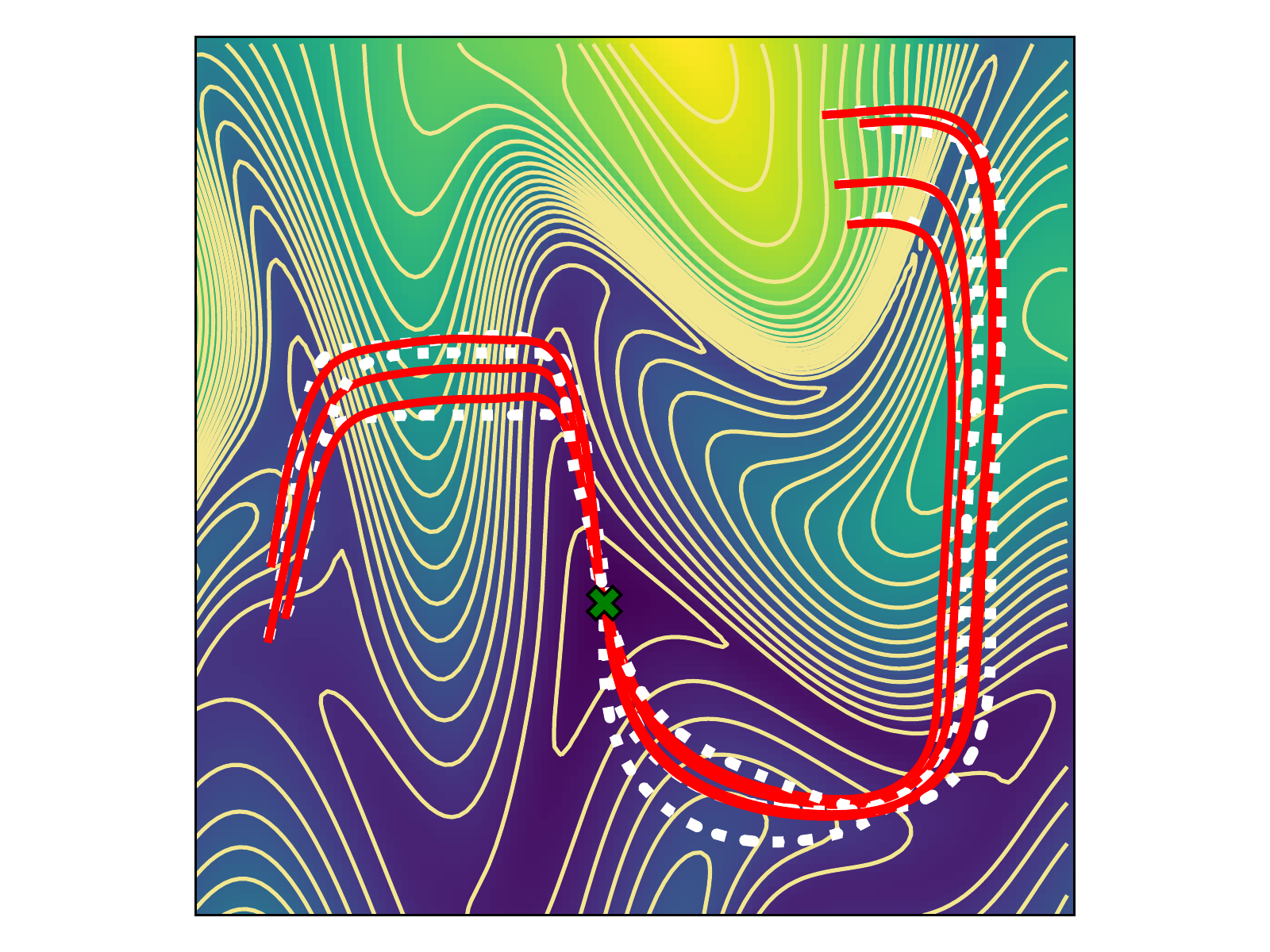}\vspace{-0.3cm}\\
\includegraphics[trim=2.5cm 0cm 2.5cm 0.2cm, clip, width = 0.16\linewidth]{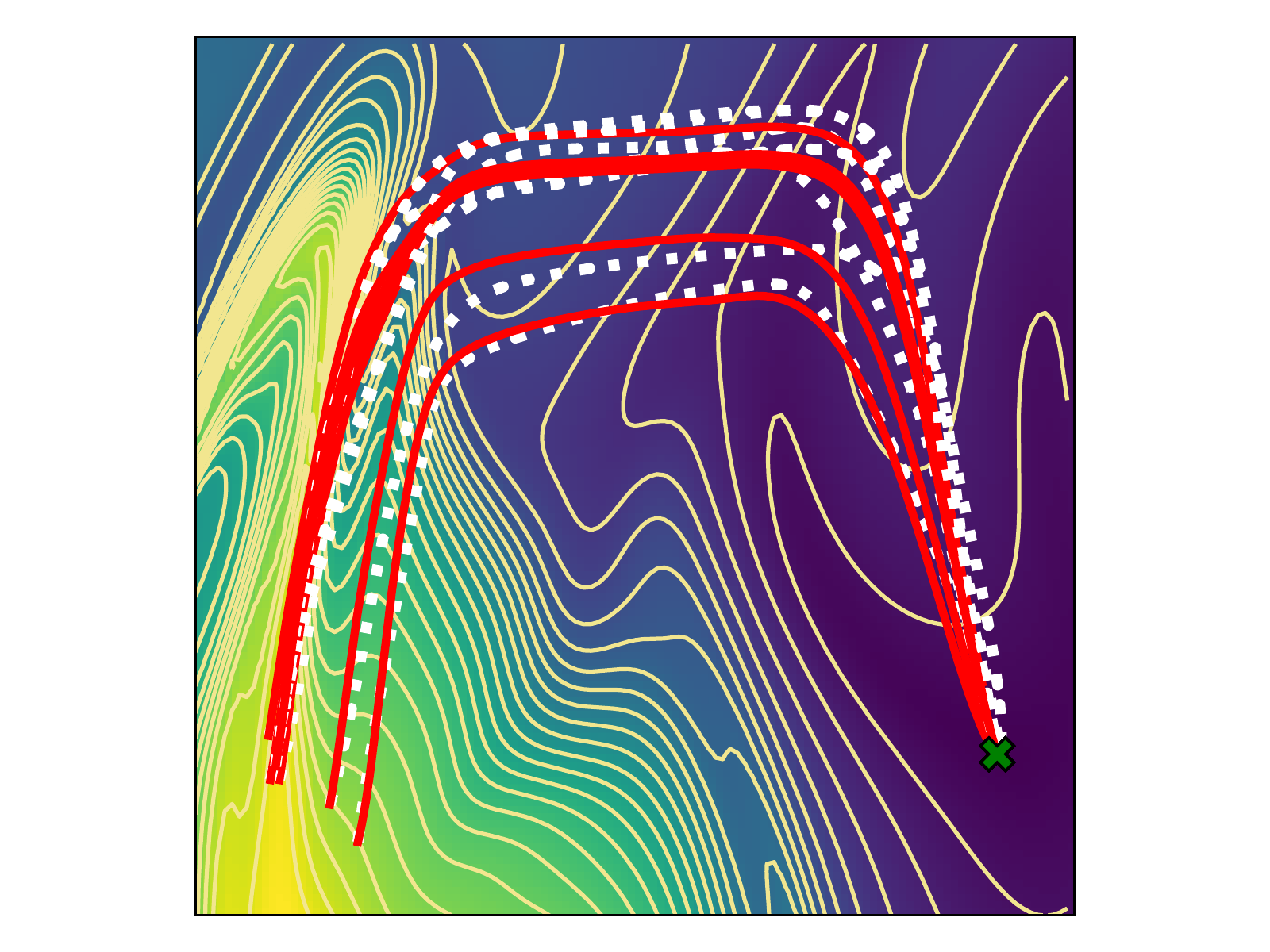}\hspace{-0.05cm}
\includegraphics[trim=2.5cm 0cm 2.5cm 0.2cm, clip, width = 0.16\linewidth]{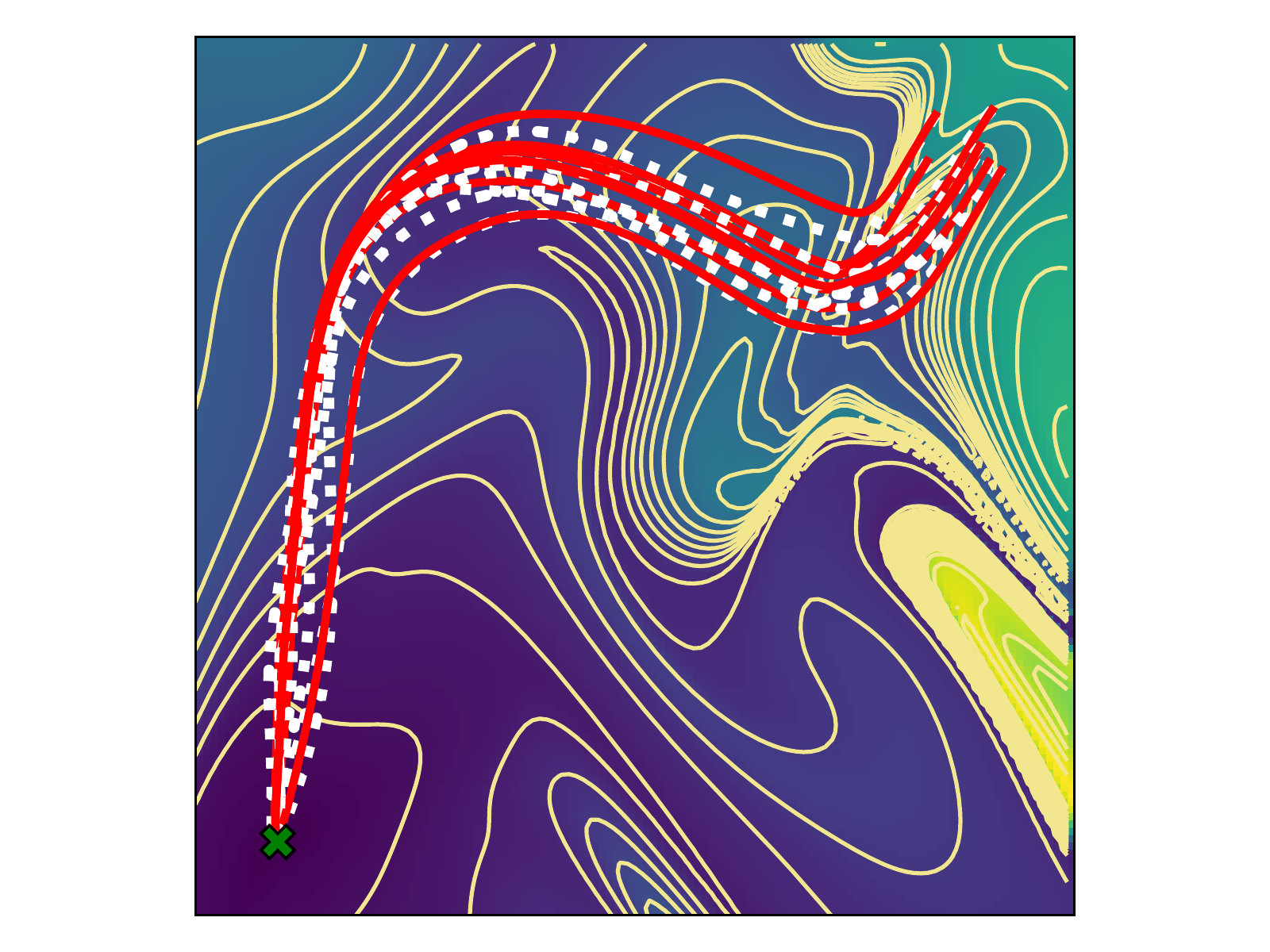}\hspace{-0.05cm} 
\includegraphics[trim=2.5cm 0cm 2.5cm 0.2cm, clip, width = 0.16\linewidth]{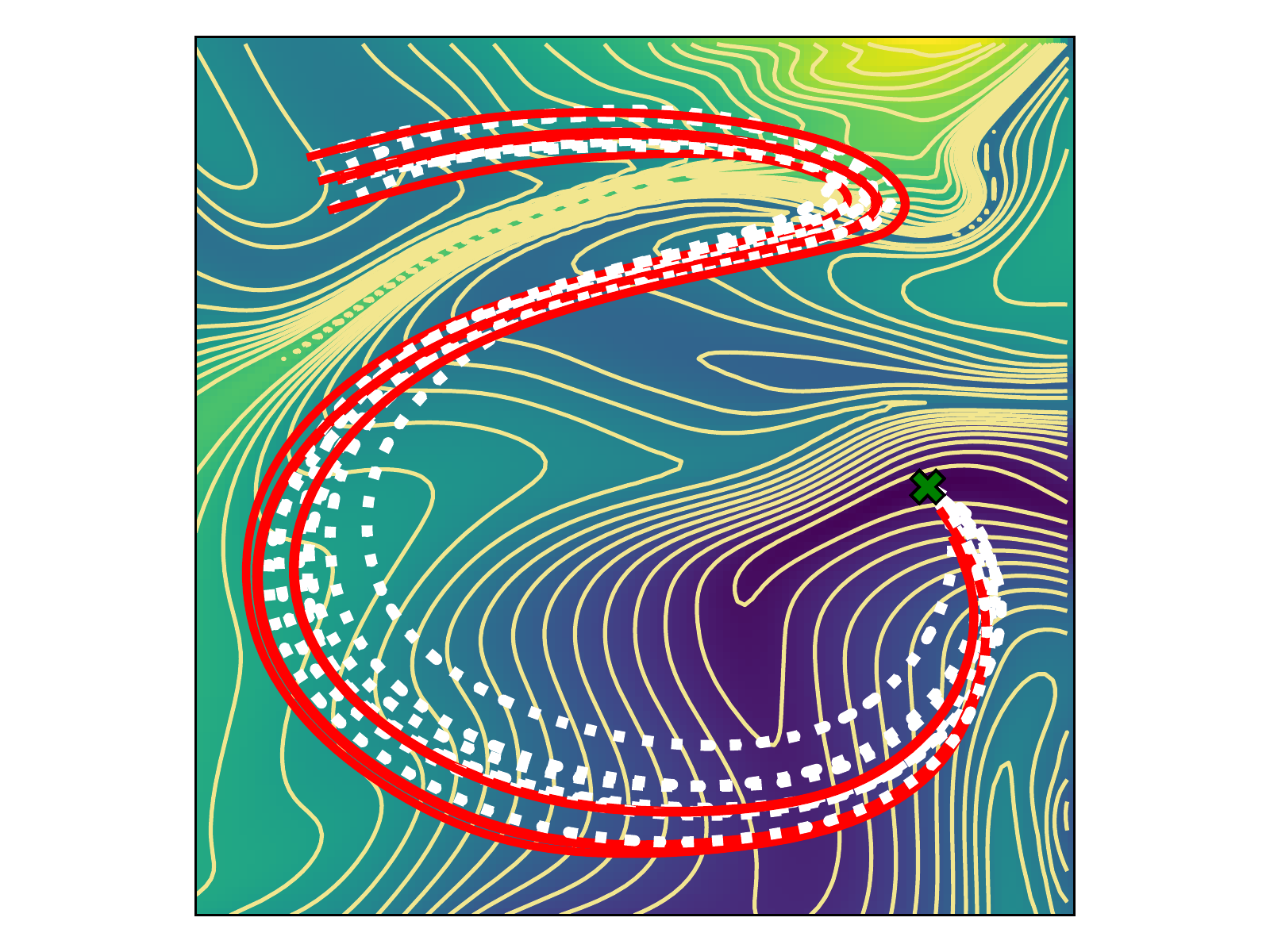}\hspace{-0.05cm} 
\includegraphics[trim=2.5cm 0cm 2.5cm 0.2cm, clip, width = 0.16\linewidth]{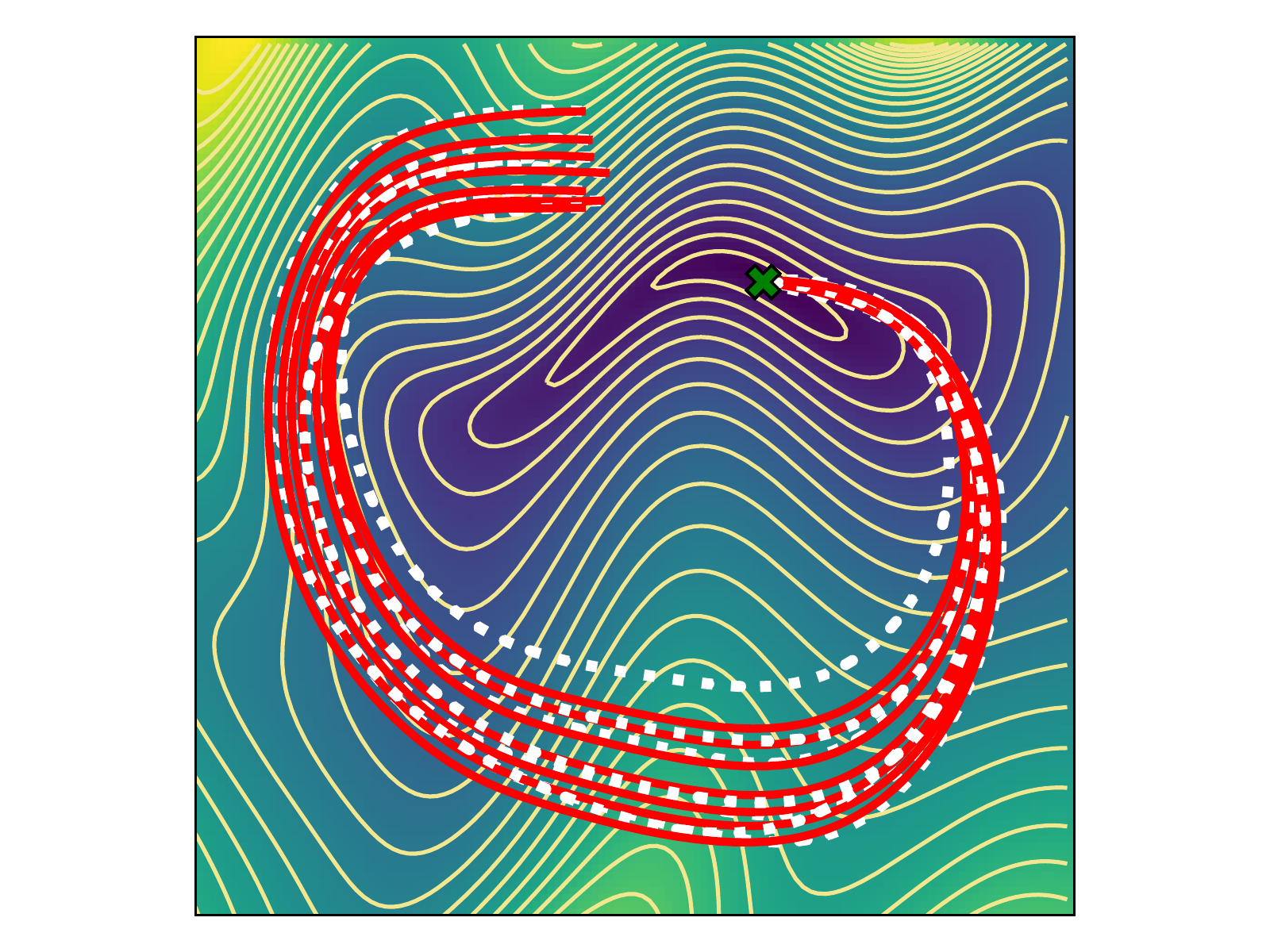}\hspace{-0.05cm}
\includegraphics[trim=2.5cm 0cm 2.5cm 0.2cm, clip, width = 0.16\linewidth]{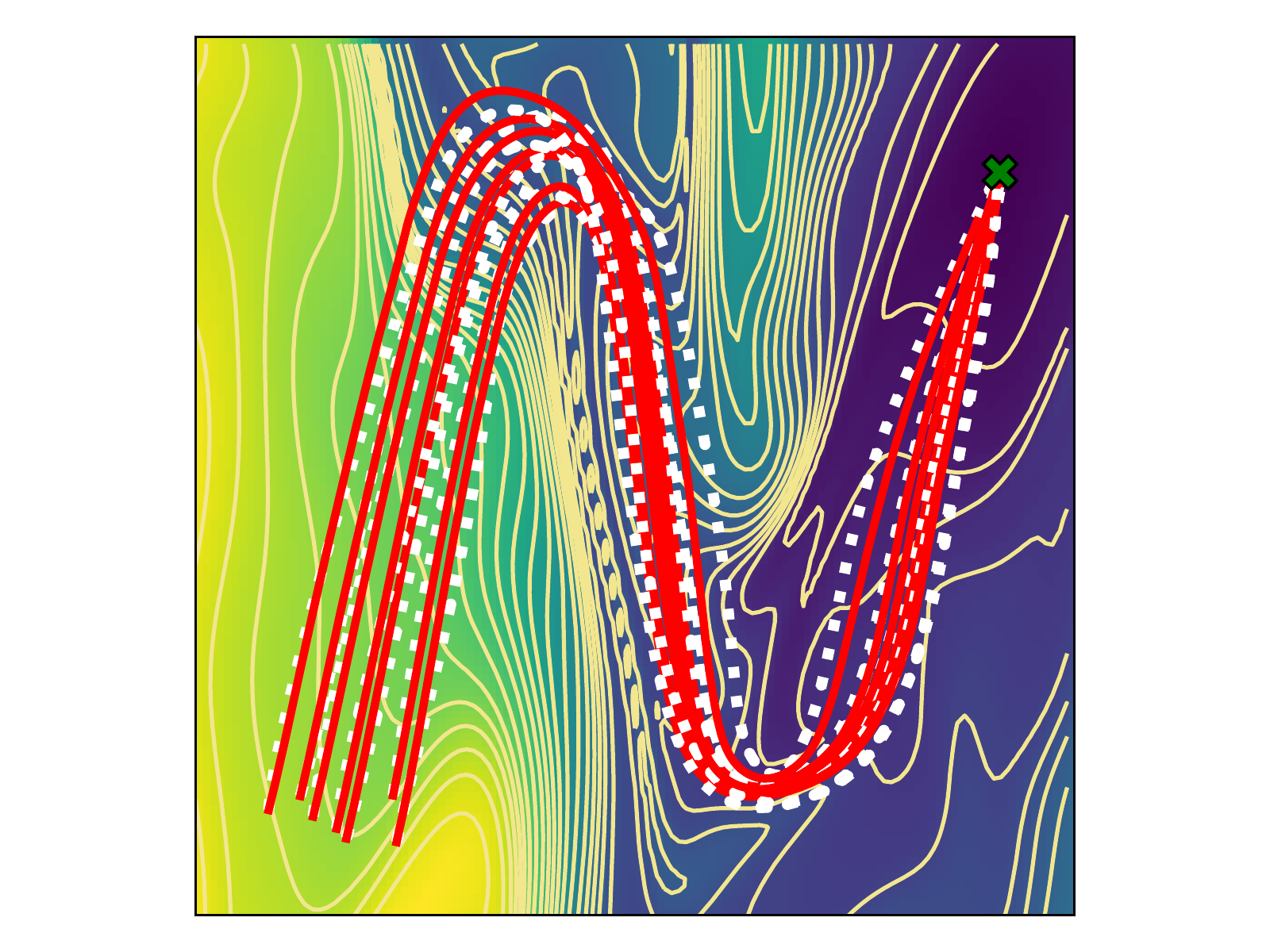}\hspace{-0.05cm}
\includegraphics[trim=2.5cm 0cm 2.5cm 0.2cm, clip, width = 0.16\linewidth]{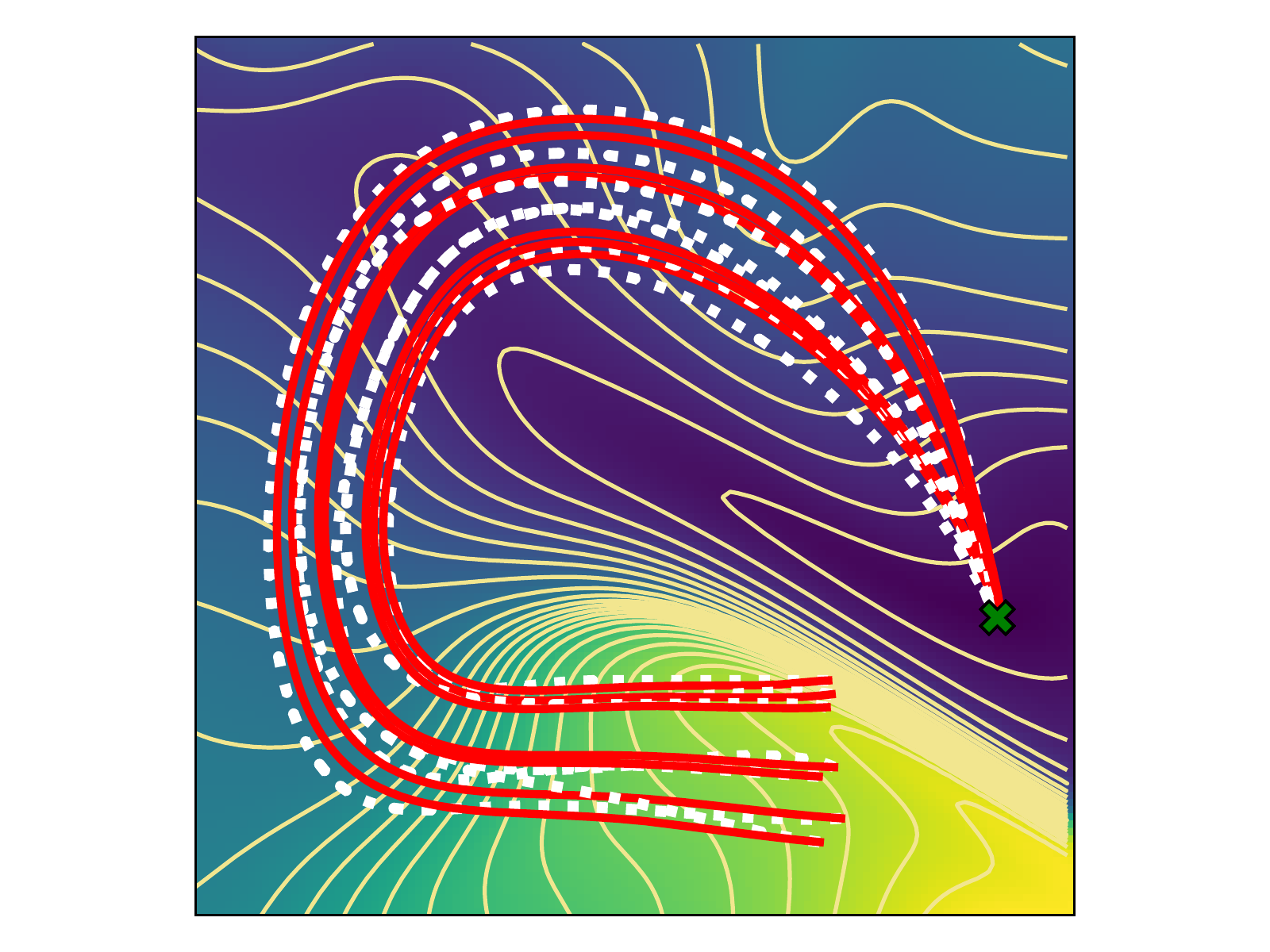}
\end{tabular}
\caption{\small{Isocontours of the potential function on LASA  dataset. Overlayed are the demonstrations (white) and the reproductions (red). The reproductions cut across the isocontours.}}
\label{fig:lasa_contour_plots}
\end{figure}

\begin{figure}[htb]
\begin{tabular}{cccccc}
\includegraphics[trim=2.5cm 0.1cm 2.5cm 0cm, clip, width = 0.16\linewidth]{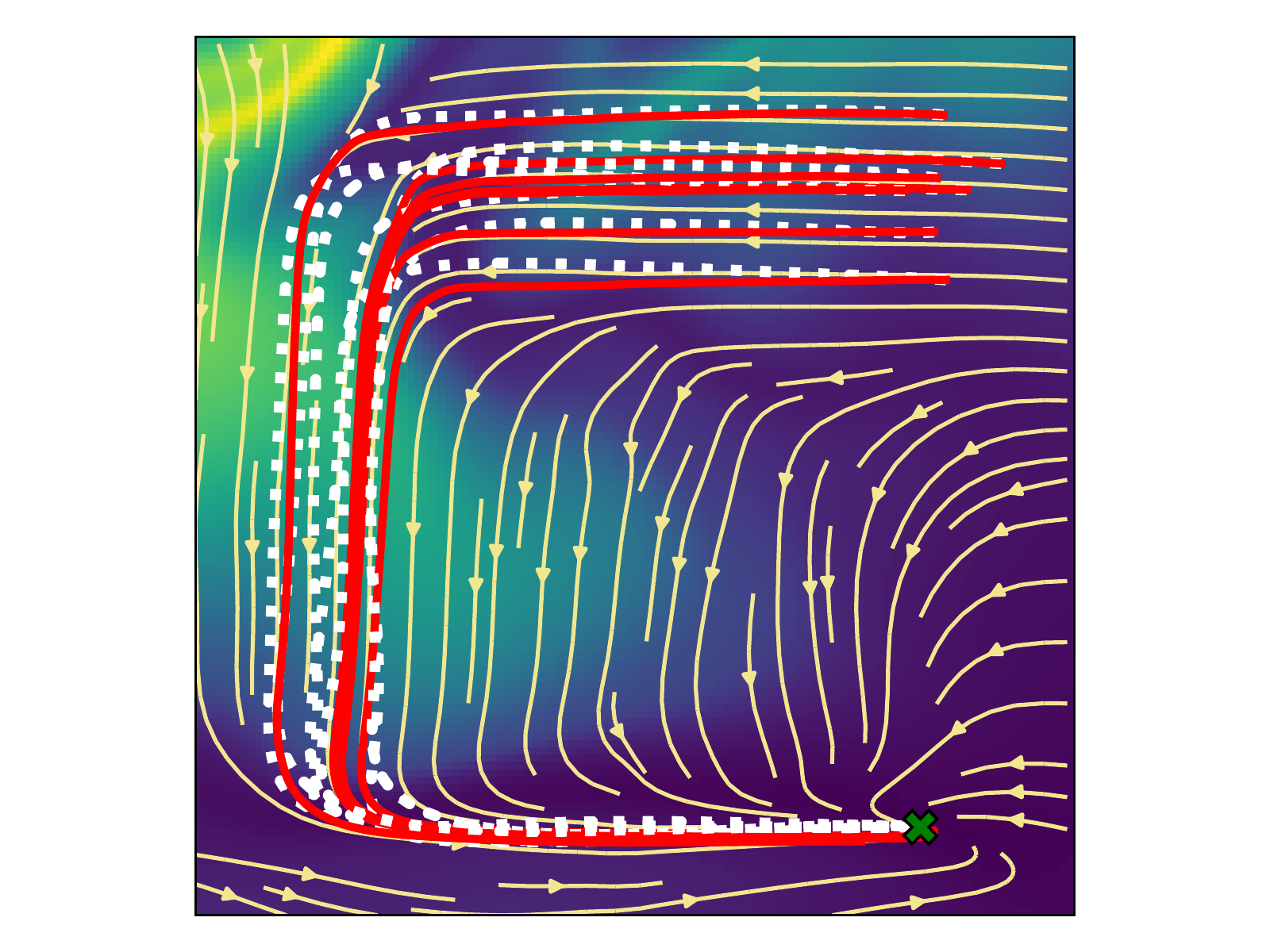}\hspace{-0.05cm}
\includegraphics[trim=2.5cm 0.1cm 2.5cm 0cm, clip, width = 0.16\linewidth]{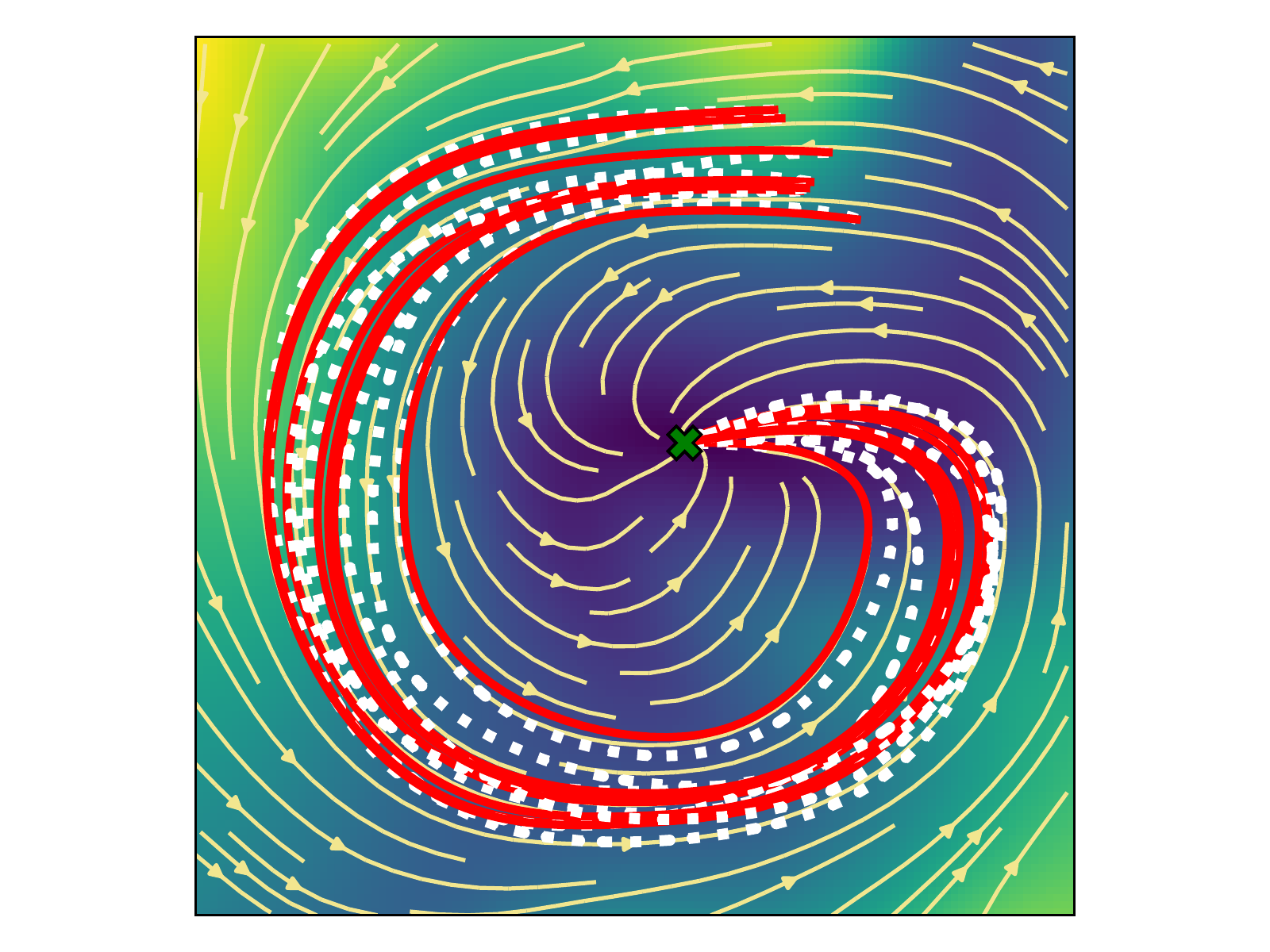}\hspace{-0.05cm}
\includegraphics[trim=2.5cm 0.1cm 2.5cm 0cm, clip, width = 0.16\linewidth]{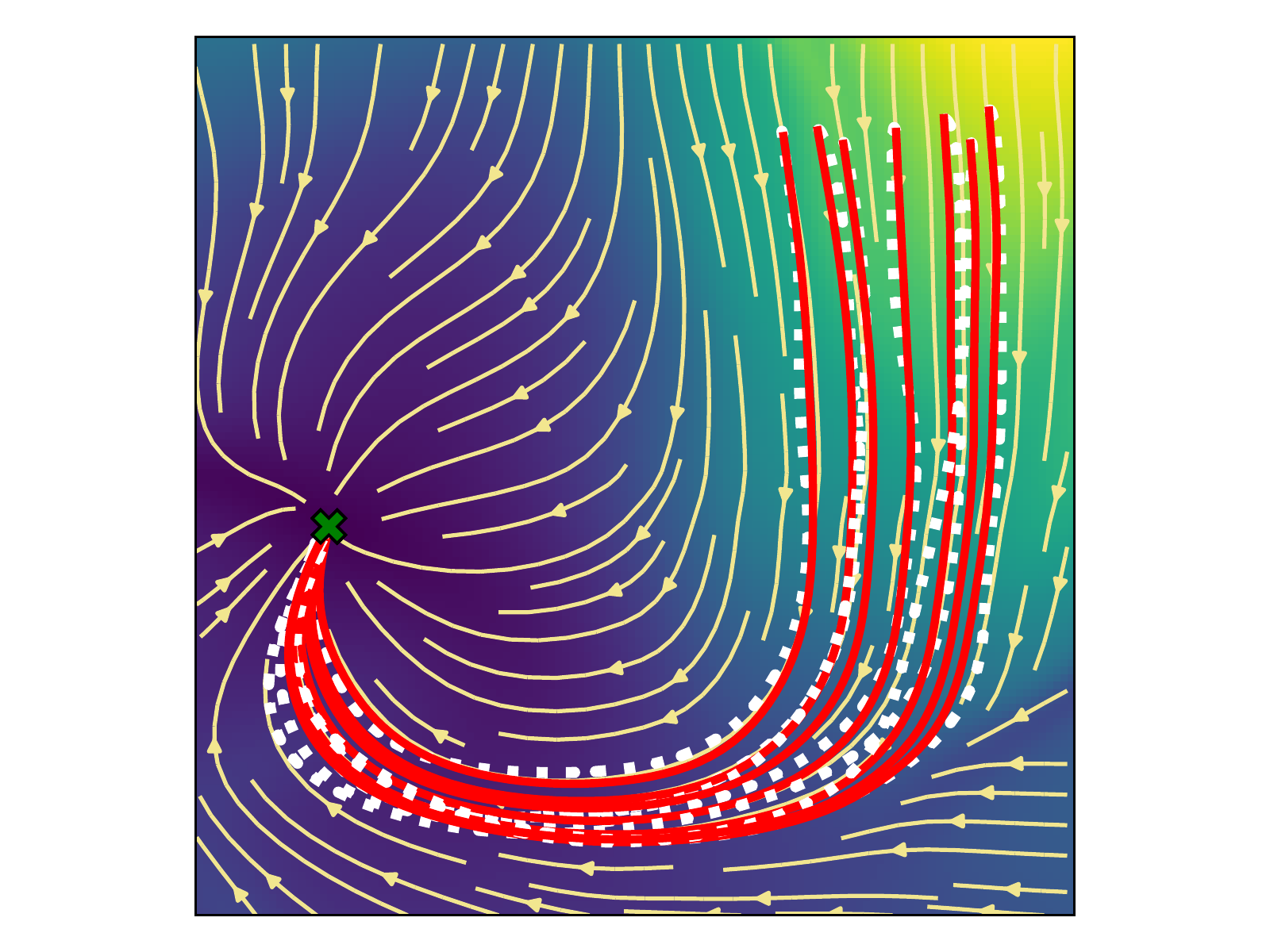}\hspace{-0.05cm} 
\includegraphics[trim=2.5cm 0.1cm 2.5cm 0cm, clip, width = 0.16\linewidth]{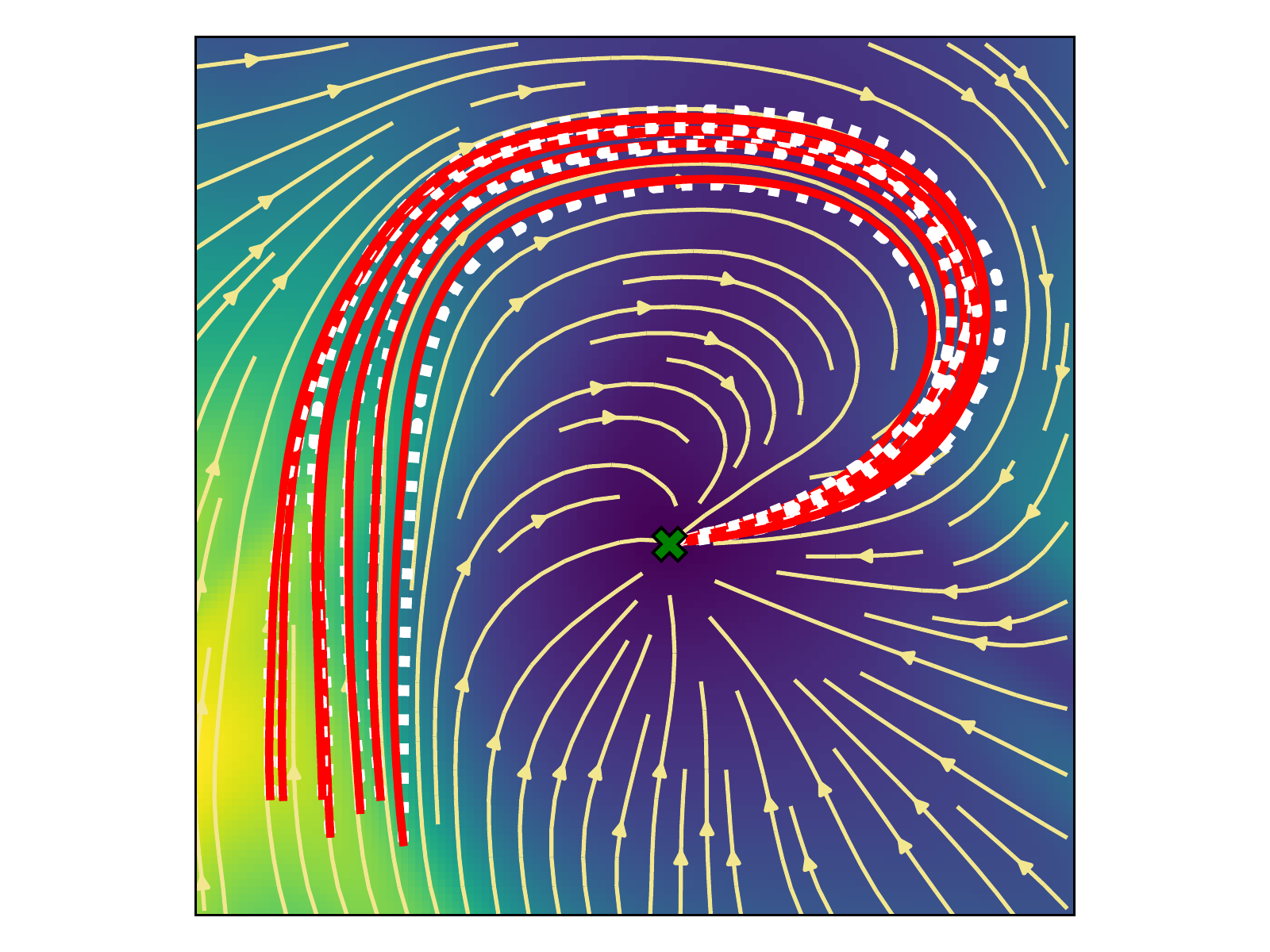}\hspace{-0.05cm}
\includegraphics[trim=2.5cm 0.1cm 2.5cm 0cm, clip, width = 0.16\linewidth]{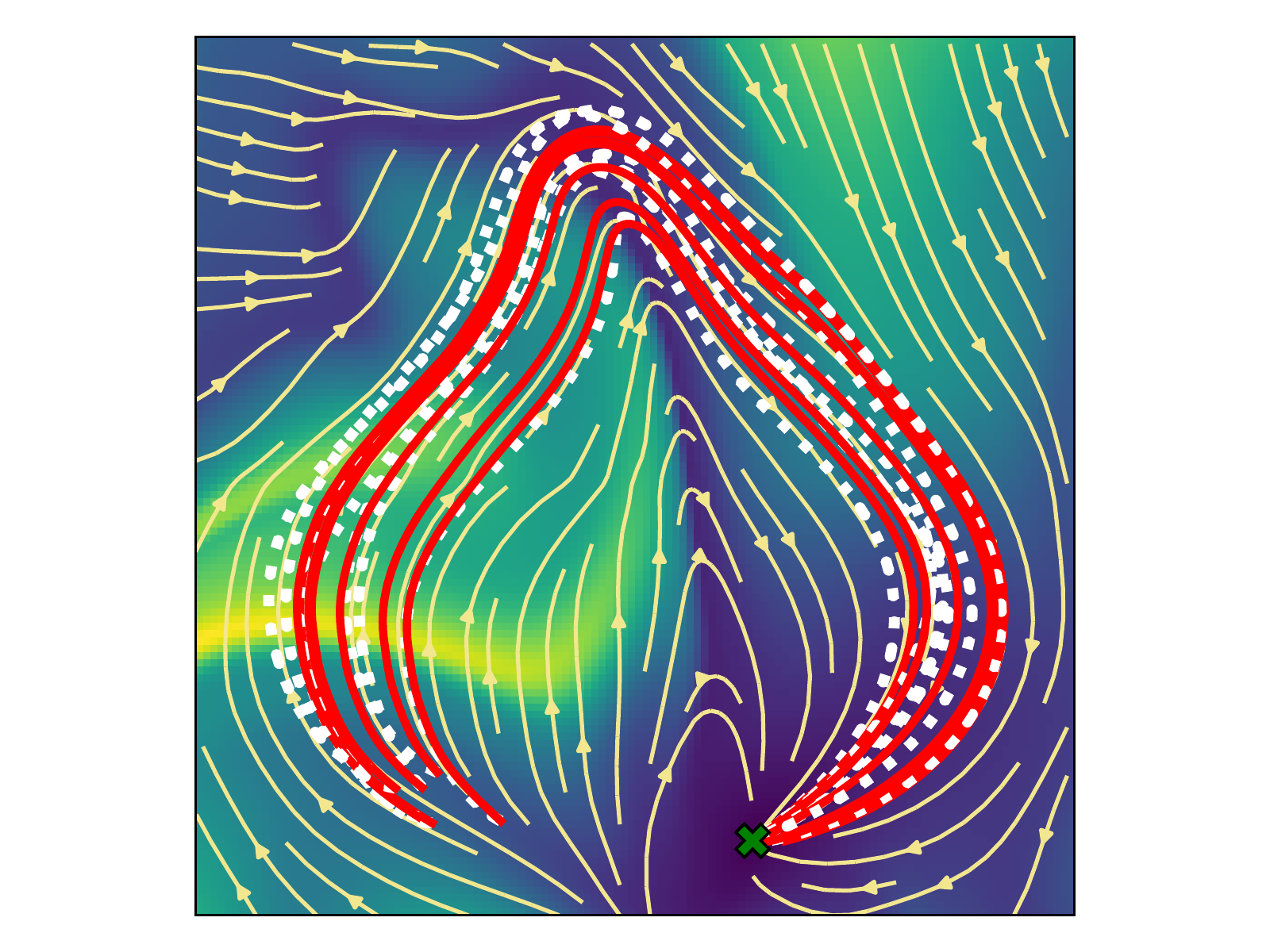}\hspace{-0.05cm}
\includegraphics[trim=2.5cm 0.1cm 2.5cm 0cm, clip, width = 0.16\linewidth]{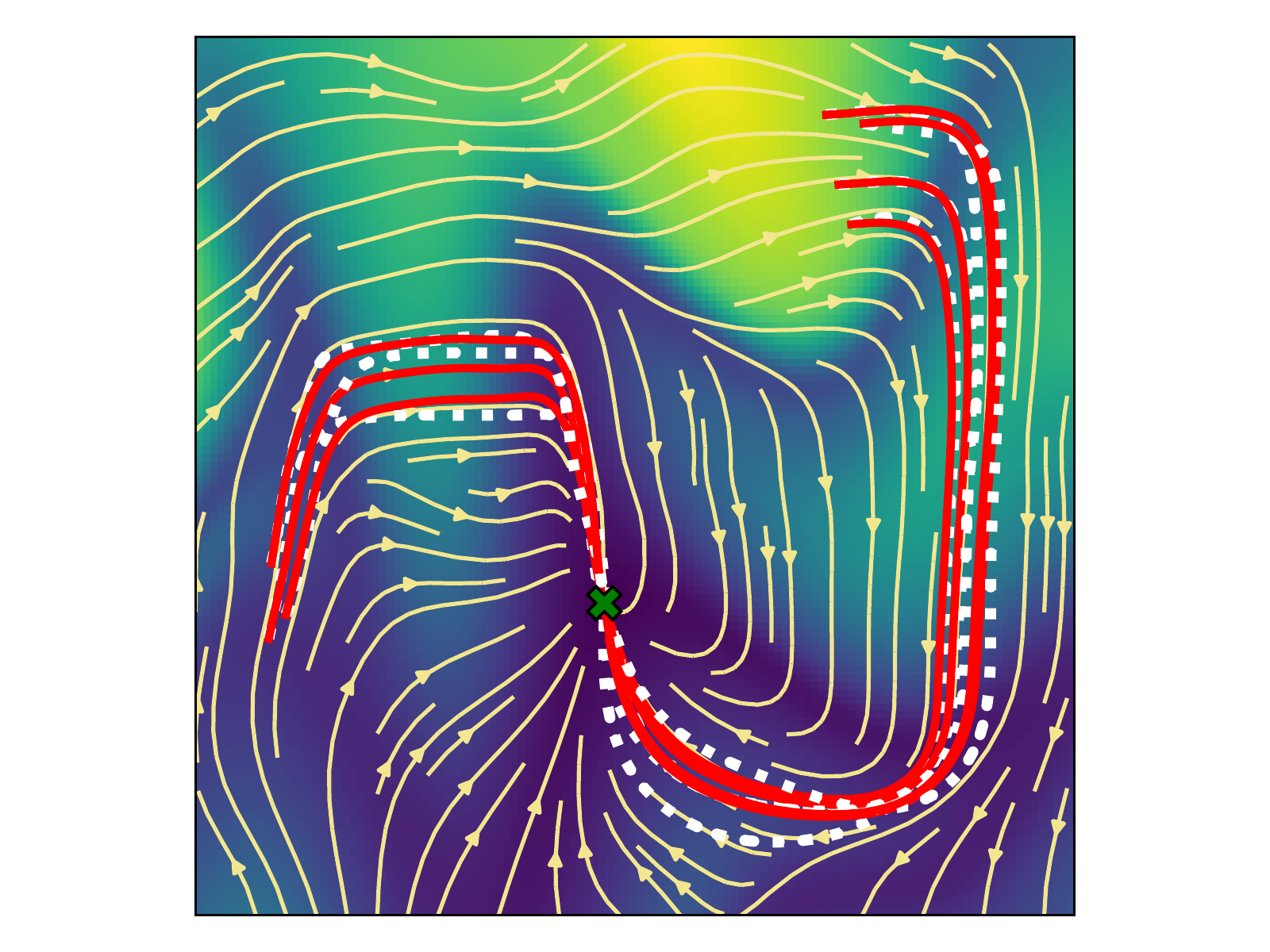}\vspace{-0.3cm}\\
\includegraphics[trim=2.5cm 0cm 2.5cm 0.2cm, clip, width = 0.16\linewidth]{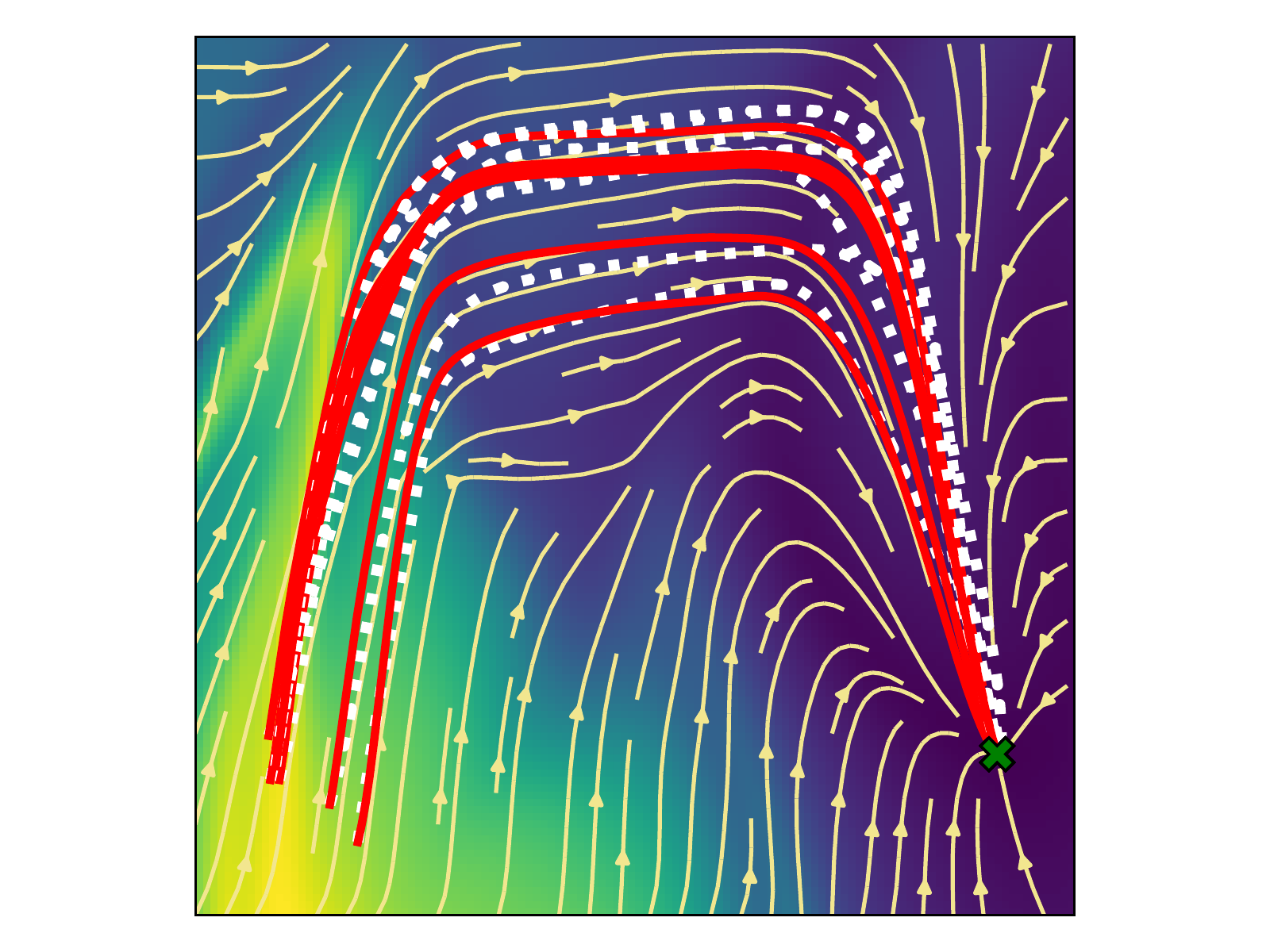}\hspace{-0.05cm}
\includegraphics[trim=2.5cm 0cm 2.5cm 0.2cm, clip, width = 0.16\linewidth]{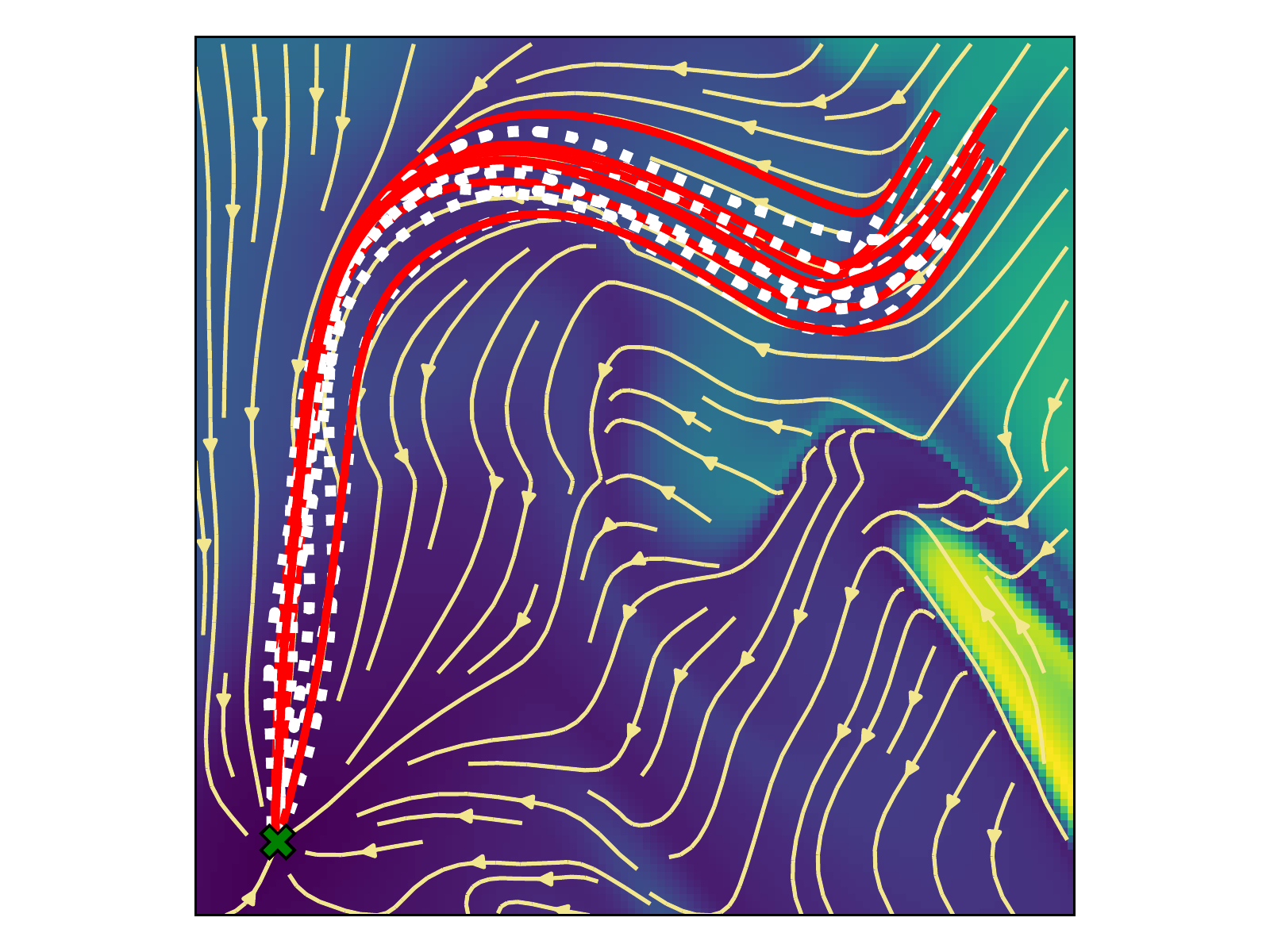}\hspace{-0.05cm} 
\includegraphics[trim=2.5cm 0cm 2.5cm 0.2cm, clip, width = 0.16\linewidth]{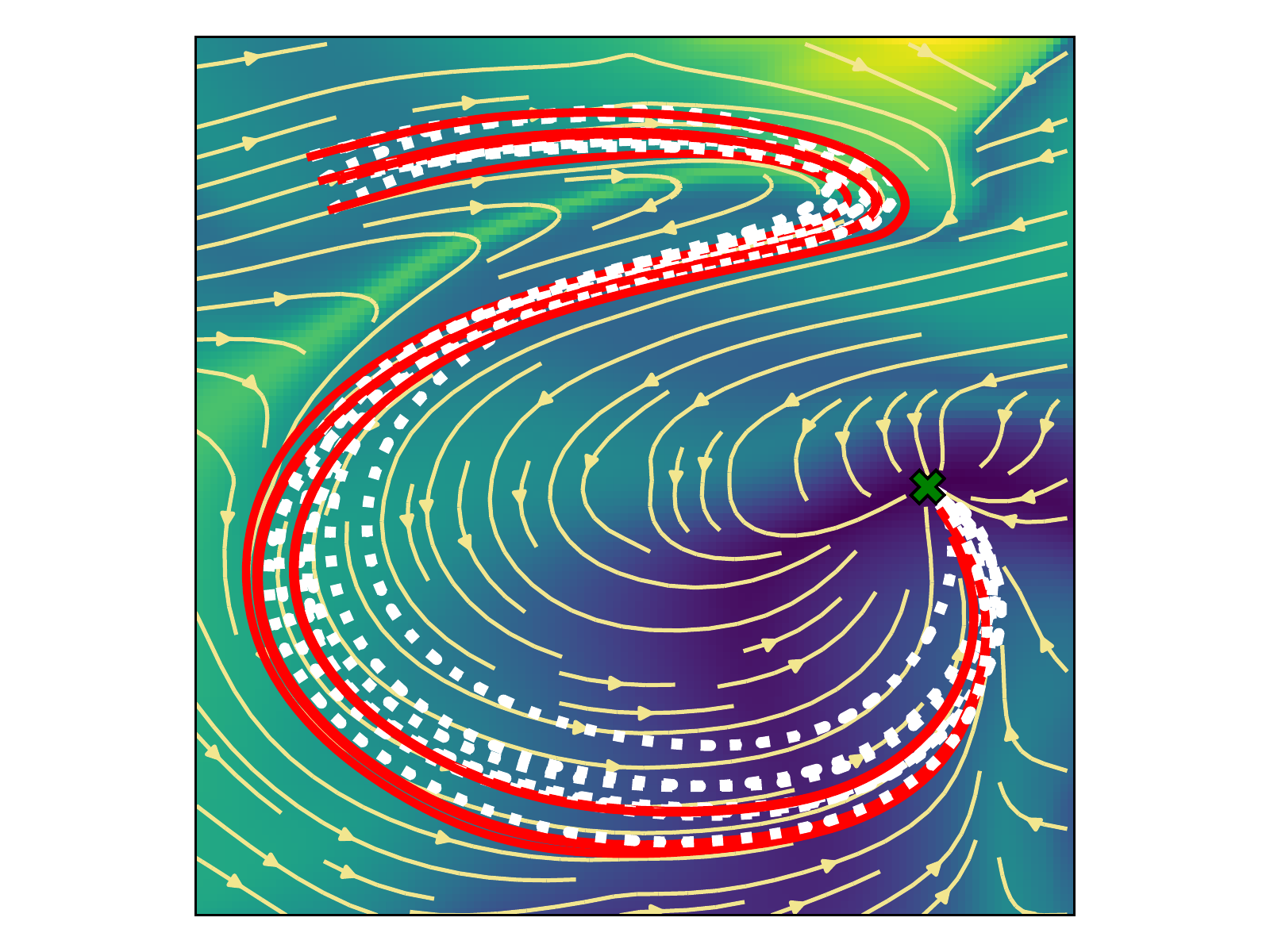}\hspace{-0.05cm} 
\includegraphics[trim=2.5cm 0cm 2.5cm 0.2cm, clip, width = 0.16\linewidth]{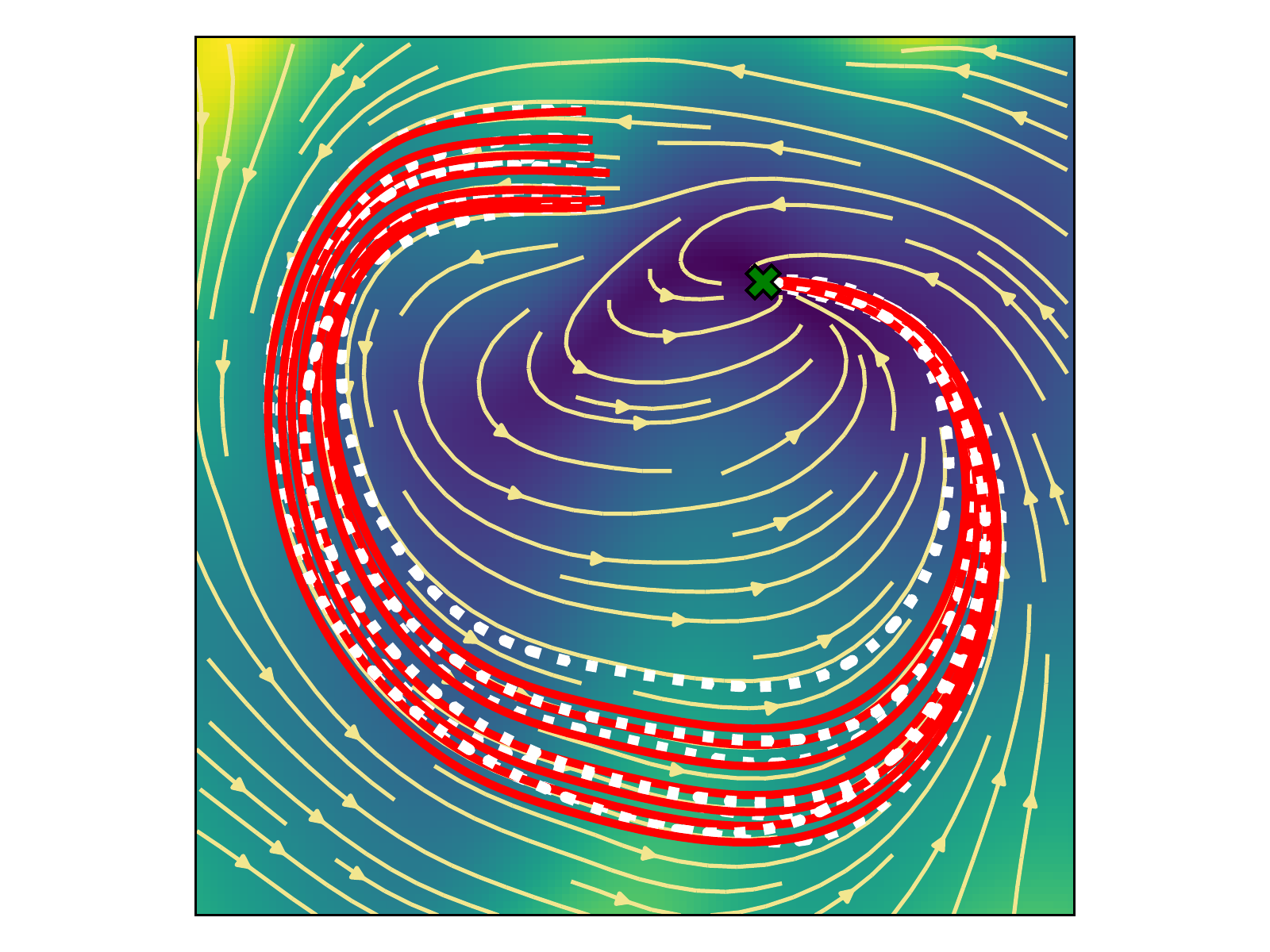}\hspace{-0.05cm}
\includegraphics[trim=2.5cm 0cm 2.5cm 0.2cm, clip, width = 0.16\linewidth]{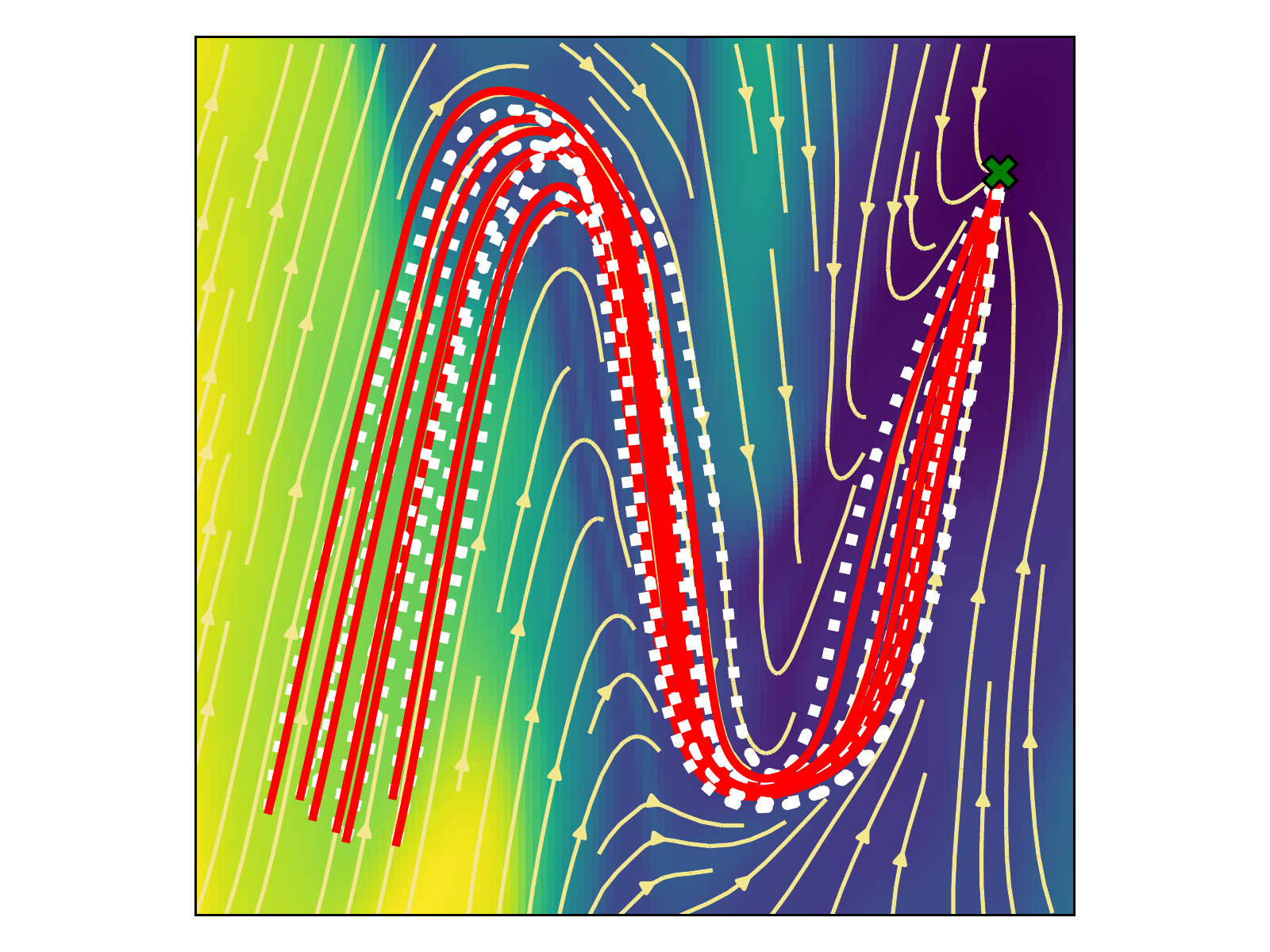}\hspace{-0.05cm}
\includegraphics[trim=2.5cm 0cm 2.5cm 0.2cm, clip, width = 0.16\linewidth]{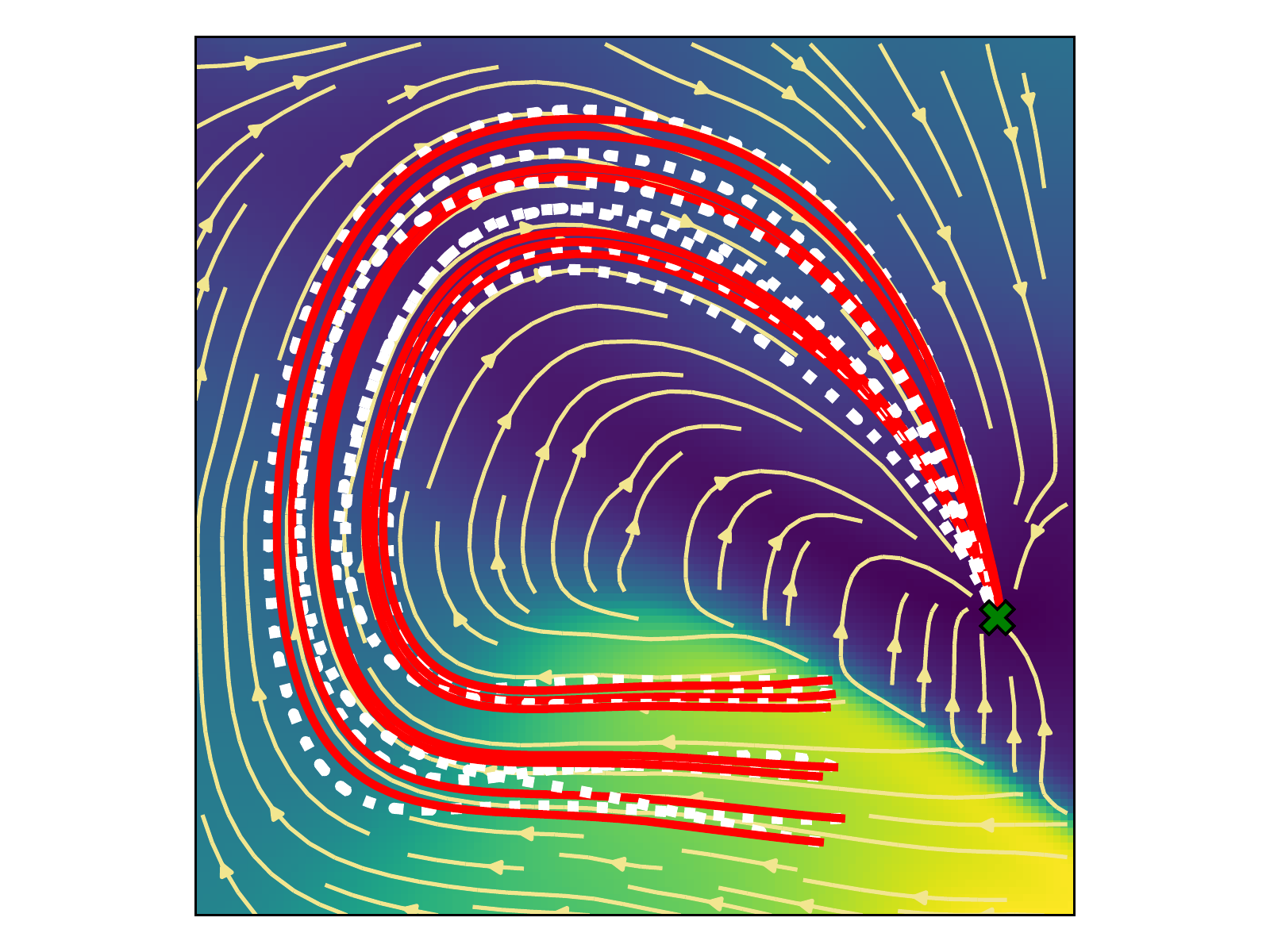}
\end{tabular}
\caption{\small{Vector fields of the dynamics learned on the LASA dataset, alongside demonstrations (white) and reproductions (red). The reproductions are governed by the natural gradient descent dynamics.}}
\label{fig:lasa_vector_fields}
\end{figure}

The LASA dataset~\cite{khansari2019lasa} consists of a library of 30 two-dimensional handwritten letters, each with 7 demonstrations. The scale of the dataset is $100mm \times 100mm$. For each letter, we find a diffeomorphism using \eqref{eq:learning_prob}. Fig. \ref{fig:lasa_vector_fields} shows the vector fields as governed by \eqref{eq:natural_gd} on a subset of letters.\begin{wrapfigure}{r}{0.28\textwidth}
    \centering
    \vspace{-4mm}
    \includegraphics[width=0.28\textwidth]{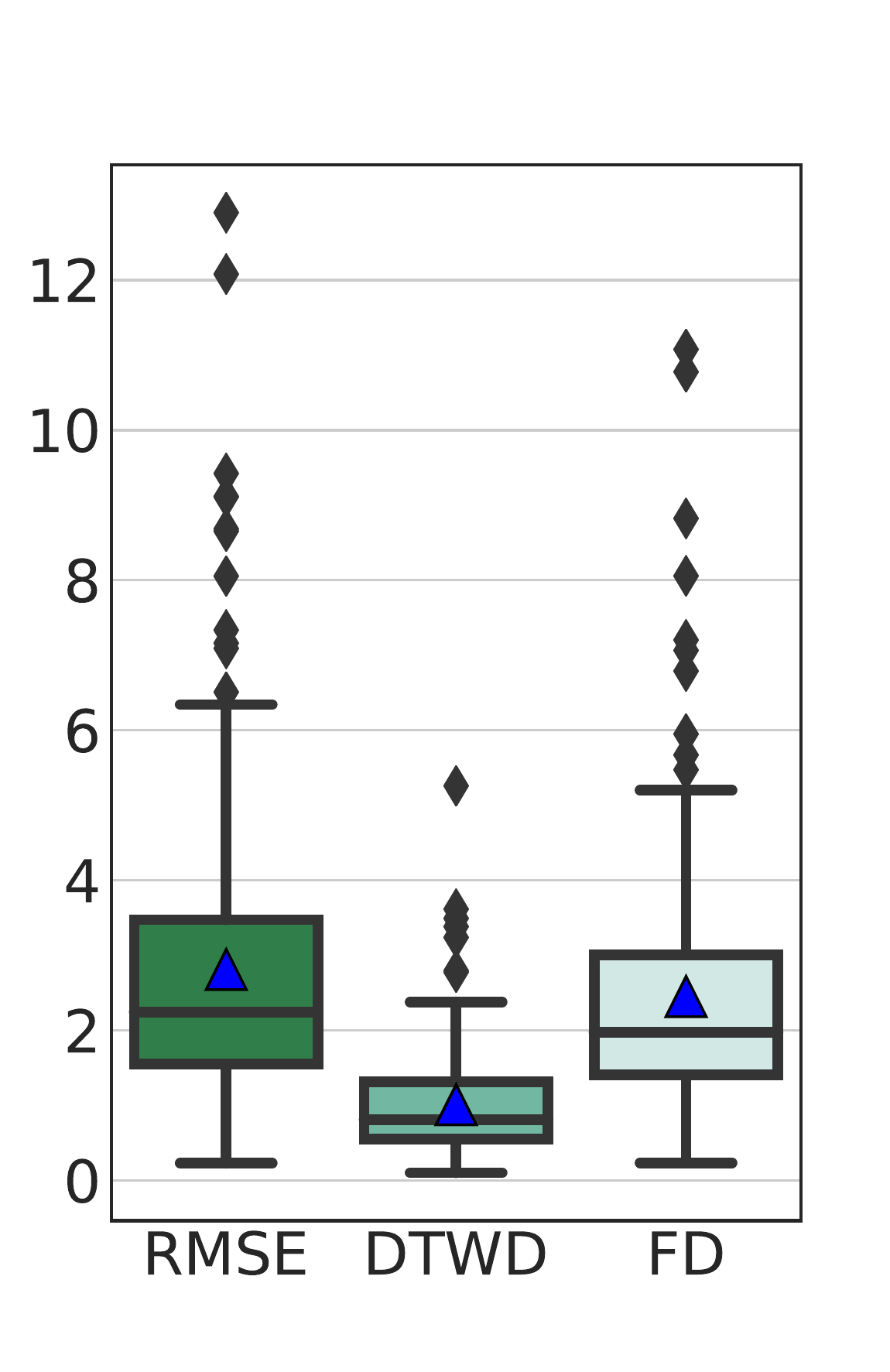}
    \caption{\small{Box plots for RMSE, average DTWD, and FD, in millimeters, evaluated over the LASA dataset. The overlayed blue triangles are the means for each metric.}}
    \vspace{-4mm}
    \label{fig:box_plot}
\end{wrapfigure} In all the plots, the rollouts (in red) closely match the demonstrations (in white), coming to rest at the goal. Furthermore, due to the structure imposed in learning, the dynamical system generalizes smooth and stable motions throughout the state-space. Also shown in Fig. \ref{fig:lasa_contour_plots} are isocontours of the potential function $\Phi({\psi}(x))$. For quantitative evaluations, we employ three error metrics: \emph{root mean squared error (RMSE)}, \emph{dynamic time warping distance (DTWD)}~\cite{berndt1994using}, and \emph{Frechet disance (FD)}~\cite{frechet1906quelques}. These metrics evaluate performance of our approach in terms of its capability to reproduce the demonstrated motions. Fig. \ref{fig:box_plot} reports these metrics, in millimeters, evaluated over $210$ demonstrations ($7$ letters $\times$ $30$ demonstrations). Each aforementioned metric focuses on different aspects of the motions. RMSE penalizes both spatial and temporal misalignment between demonstrated and reproduced motions. On the other hand, DTWD and FD disregard time misalignment, and instead focus solely on the spatial misalignment between motions. Since DTWD between any two trajectories is a time-aggregated measure of error, we report \emph{average} DTWD, found by dividing the DTWD by the number of points $T_i$ in a trajectory. In Fig.~\ref{fig:box_plot}, the median and mean errors are observed to be small relative to the scale of the data, signifying that the learned dynamical systems are able to accurately reproduce most motions. However, higher errors are occasionally observed, accounting for outliers. This is mostly due to intersecting demonstrations which can not be modeled by a first-order dynamical system. 

\begin{figure}[htb!]
    \centering
    \fbox{
    \includegraphics[trim=2.4cm 5cm 10cm 3cm, clip,width=0.20\textwidth,valign=c]{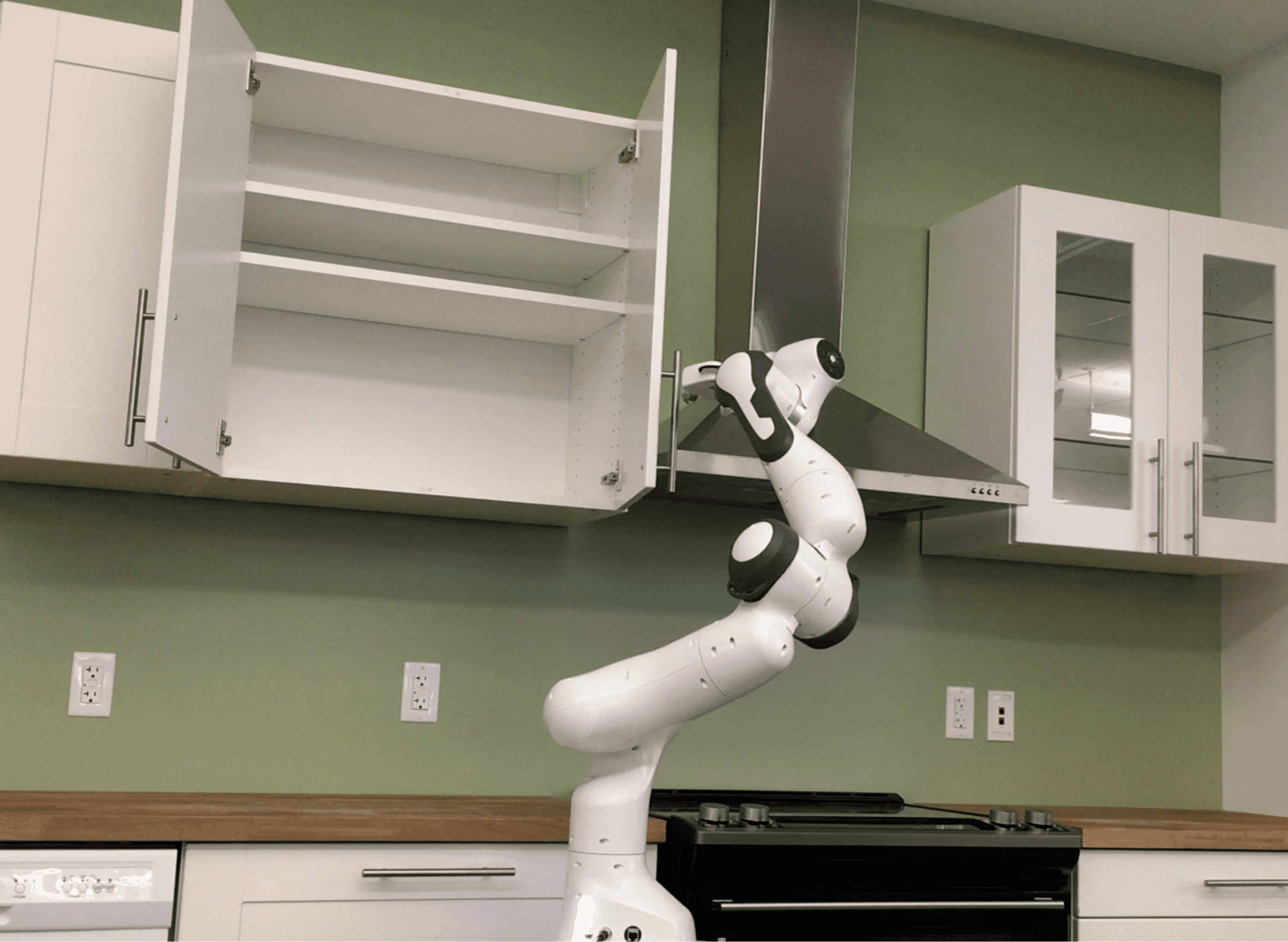}
    \includegraphics[trim=2.5cm 0.2cm 2.5cm 0.2cm, clip,width=0.20\textwidth,valign=c]{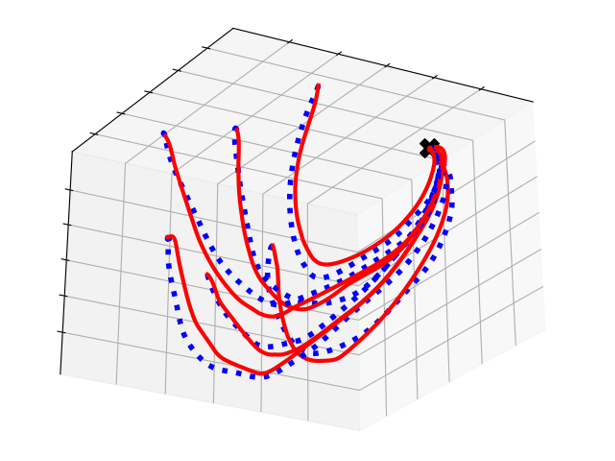}
    }
    ~
    \fbox{
    \includegraphics[trim=2.4cm 5cm 10cm 3cm, clip,width=0.20\textwidth,valign=c]{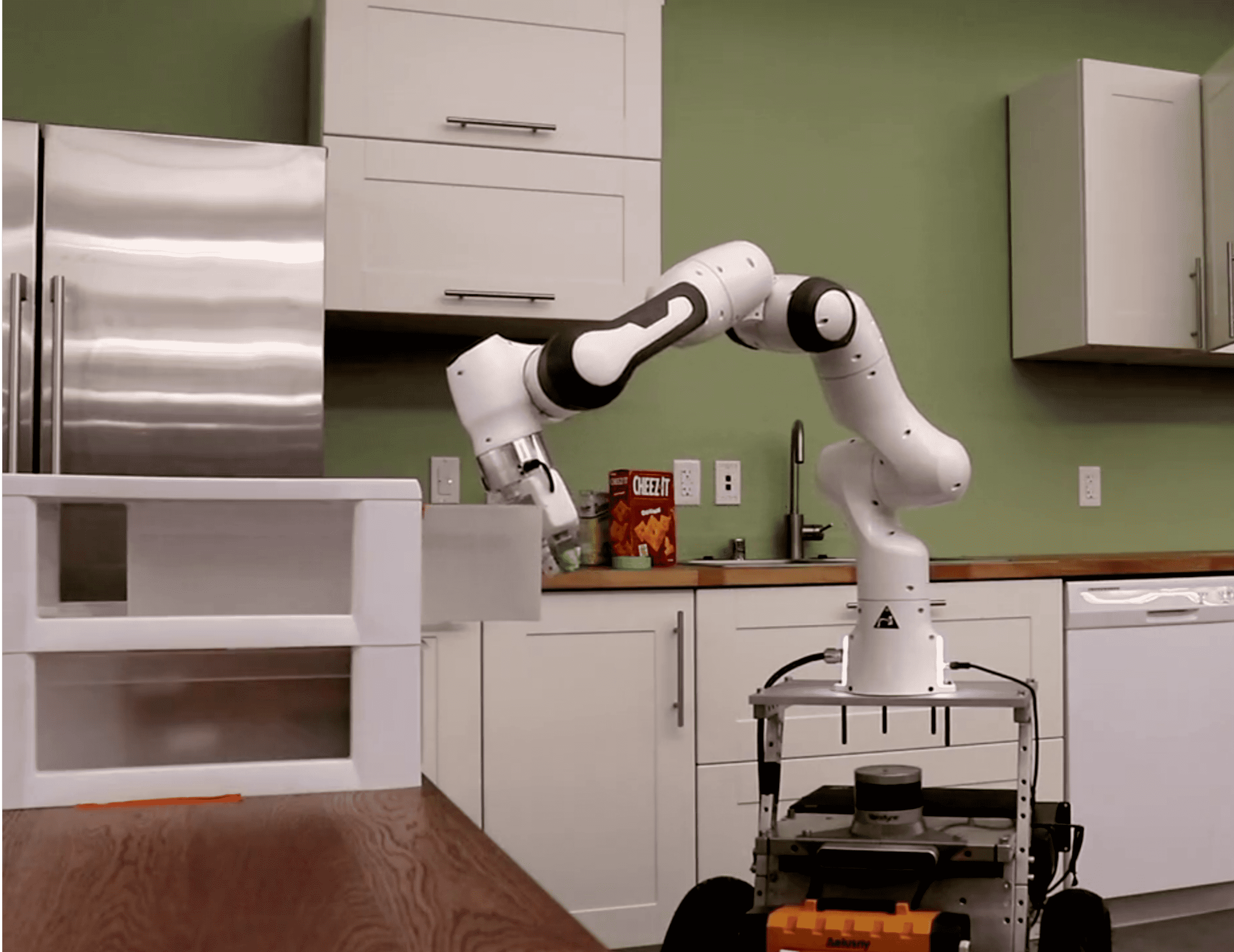}
    \includegraphics[trim=2.5cm 0.2cm 2.5cm 0.2cm, clip,width=0.20\textwidth,valign=c]{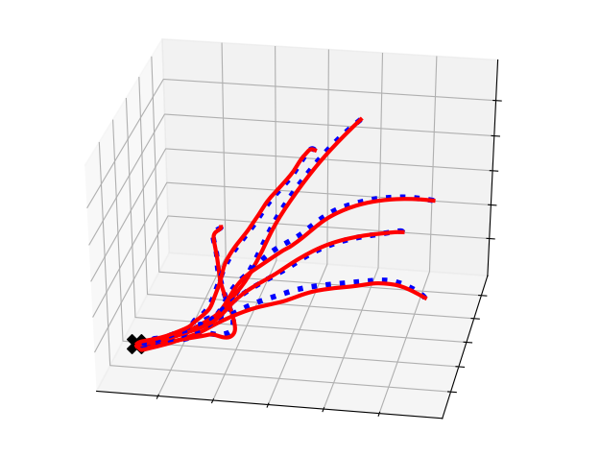}
    }
    \vspace{2mm}
    \caption{\small{The \emph{door reaching} (left) and \emph{drawer closing tasks} (right). The demonstrations are plotted in blue while the reproductions are in red.}}
    \vspace{4mm}
    \label{fig:franka_experiments}
\end{figure}

For demonstrations collected on a Franka Emika robot, we evaluate on two tasks:  \emph{door reaching}, and \emph{drawer closing}~\cite{rana2019learning}. Each task dataset consists of 6 three-dimensional end-effector motions collected by physically guiding the robot. The \emph{door reaching} task required the robot to start from inside a cabinet and reach the door handle, while the \emph{drawer closing} task required reaching a drawer handle and pushing the drawer close. Fig. \ref{fig:franka_experiments} shows the tasks, demonstrated motions (blue), as well as the reproduced motions (red). Our learning approach is observed to accurately reproduce three-dimensional motions collected on a real robot.  


\vspace{-0.5em}
\section{Conclusion}
\vspace{-0.5em}
We have presented \algo, an approach for learning motion skills from a few human demonstrations. \algo encodes complex human motions as generated from a dynamical system, linked under a learnable diffeomorphism, to a simple gradient-descent dynamical system on a latent space. A class of parameterized diffeomorphisms is proposed to learn a wide range of motions and generalize to different tasks with minimal parameter tuning. Experimental validation on a handwriting dataset and data collected on a real robot is provided to show the efficacy of the proposed approach.

\acks{This work was supported in part by NVIDIA Research.}

\bibliography{refs}

\end{document}